\documentclass{lmcs}
\pdfoutput=1

\usepackage{lastpage}
\lmcsdoi{21}{3}{16}
\lmcsheading{}{\pageref{LastPage}}{}{}%
{Aug.~12,~2024}{Aug.~07,~2025}{}

\usepackage{amsmath,amssymb,amsfonts}%
\usepackage{tikz}
\usetikzlibrary{positioning,patterns}
\usepackage{xspace}
\usepackage{calc}
\usepackage{listings}%
\usepackage{enumitem}
\usepackage{multirow}%
\usepackage{soul}
\usepackage{pagecolor}
\usepackage{color}
\usepackage[most]{tcolorbox}
\usepackage{caption}
\usepackage{algorithm}%
\usepackage{algpseudocode}%
\usepackage{mathtools}
\usepackage{booktabs}%
\usepackage{tabularx}
\usepackage{bookmark}
\usepackage{url}
\usepackage[T1]{fontenc} 

\usepackage{graphicx}%
\usepackage{amsthm}%
\usepackage{mathrsfs}%
\usepackage[title]{appendix}%
\usepackage{xcolor}%
\usepackage{textcomp}%
\usepackage{manyfoot}%
\usepackage{algorithmicx}%
\usepackage[utf8]{inputenc} 

\DeclarePairedDelimiter\ceil{\lceil}{\rceil}

\newcommand{\clingo}{\emph{clingo}\xspace}
\newcommand{\clingodl}{\emph{clingo}[DL]\xspace}
\newcommand{\figspace}{\vspace{-4pt}}

\newcommand{\stest}{\textit{ST\_EST}\xspace}
\newcommand{\stmtwr}{\textit{ST\_MTWR}\xspace}
\newcommand{\stfifo}{\textit{ST\_FIFO}\xspace}

\newcommand{\wtest}{\textit{WT\_EST}\xspace}
\newcommand{\wtmtwr}{\textit{WT\_MTWR}\xspace}
\newcommand{\wtfifo}{\textit{WT\_FIFO}\xspace}

\newcolumntype{L}[1]{>{\raggedright\arraybackslash}p{#1}}
\newcolumntype{C}[1]{>{\centering\arraybackslash}p{#1}}
\newcolumntype{R}[1]{>{\raggedleft\arraybackslash}p{#1}}

\algnewcommand{\To}{{\normalfont\bfseries to }}

\newlength{\scaledx}
\newlength{\scaledy}
\newcommand\SetScales{%
	\pgfextractx{\scaledx}{\pgfpointxy{1}{0}}%
	\pgfextracty{\scaledy}{\pgfpointxy{0}{1}}%
}

\newlength\listingnumberwidth
\urldef\myurl\url{https://github.com/prosysscience/Job-Shop-Scheduling/tree/master/ICLP\%20Paper/Implementation}

\newcommand{\redcolor}{}
\newcommand{\addarea}[1]{%
\begingroup
\redcolor
#1
\endgroup}
\newcommand{\addcolor}{\color{.}}
\newcommand{\addtext}[1]{#1}
\newcommand{\revisedtext}[1]{#1}

\sethlcolor{\thepagecolor}

\makeatletter
\def\fcapsize@figure{\normalfont\normalsize\rmfamily}
\def\fcapstyle@figure{\normalfont\normalsize\itshape}
\makeatother


\begin{document}

\title[Decomposition and Multi-shot ASP Solving for Job-shop Scheduling]{Decomposition Strategies and Multi-shot ASP Solving for Job-shop Scheduling}



\author[M.~M.~S.~El-Kholany]{Mohammed M. S. El-Kholany\lmcsorcid{0000-0002-1088-2081}}
\author[M.~Gebser]{Martin Gebser\lmcsorcid{0000-0002-8010-4752}}
\author[K.~Schekotihin]{Konstantin Schekotihin\lmcsorcid{0000-0002-0286-0958}}
\address{Cairo University, Egypt}
\email{m.saadelden@fci-cu.edu.eg}
\address{University of Klagenfurt, Austria}
\email{mohammed.el-kholany@aau.at, martin.gebser@aau.at, konstantin.schekotihin@aau.at}







\begin{abstract}
The Job-shop Scheduling Problem (JSP) is a well-known and challenging combinatorial optimization problem in which tasks sharing a machine are to be arranged in a sequence such that encompassing jobs can be completed as early as possible. In this paper, we investigate problem decomposition into time windows whose operations can be successively scheduled and optimized by means of multi-shot Answer Set Programming (ASP) solving. From a computational perspective, decomposition aims to split highly complex scheduling tasks into better manageable subproblems with a balanced number of operations such that good-quality or even optimal partial solutions can be reliably found in a small fraction of runtime. We devise and investigate a variety of decomposition strategies in terms of the number and size of time windows as well as heuristics for choosing their operations. Moreover, we incorporate time window overlapping and compression techniques into the iterative scheduling process to counteract optimization limitations due to the restriction to window-wise partial schedules. Our experiments on different JSP benchmark sets show that successive optimization by multi-shot ASP solving leads to substantially better schedules within tight runtime limits than single-shot optimization on the full problem. In particular, we find that decomposing initial solutions obtained with proficient heuristic methods into time windows leads to improved solution quality.
\end{abstract}



\maketitle

\section{Introduction}\label{Introduction}

Effective scheduling methods are essential for complex manufacturing and transportation systems, where allocating and performing diverse tasks within resource capacity limits is one of the most critical challenges for production management \cite{uzsoy2000performance}.
The Job-shop Scheduling Problem (JSP) \cite{baker1974introduction,taillard1993benchmarks}
constitutes a well-known mathematical abstraction of industrial production scheduling in
which 
operations need to be processed by machines such that a given objective,
like the makespan for completing all jobs or their tardiness w.r.t.\ deadlines, is minimized.
Finding optimal JSP solutions, determined by a sequence of operations for each machine, is an NP-hard combinatorial problem \cite{garey1976complexity,lenstra1977complexity,liu2008prediction}. Therefore, optimal schedules and termination guarantees can be extremely challenging or even unreachable for complete optimization methods, already for moderately sized instances.
For example, it took about $20$ years 
to find a (provably) optimal solution for an instance called FT10 with $10$ jobs
\cite{adams1988shifting,zhang2010hybrid}, each consisting of a sequence of $10$ operations to be processed by $10$ machines. 

\addarea{%
Roughly speaking, the optimization approaches to JSP can be classified into exact and approximation methods \cite{xishrehu22a}.
The former guarantee optimal schedules upon termination of their complete
optimization algorithms, while the latter aim at finding good-quality
schedules without exhaustively traversing the search space.
In view of the combinatorial explosion faced when the instance size grows,
exact methods manage to guarantee optimal solutions in limited time for
reasonably small or extraordinarily simple JSP instances only.
This makes approximation methods attractive whenever termination guarantees
are beyond reach and the optimization task turns into finding the best
possible solution within a tight runtime limit.
Considering those instances in Taillard's benchmark set
\cite{taillard1993benchmarks} whose optima are known,
the optimal schedules were actually found by approximation methods,
and theoretical lower bounds rather than exhaustive search could be applied
to exclude the existence of any better solution.}

In this paper, we combine and significantly extend our studies \cite{el2022problem,elscge22a} 
on problem decomposition into time windows and successive schedule optimization
through an extension of Answer Set Programming (ASP) \cite{lifschitz19a}
with Difference Logic (DL) constraints \cite{gebser2016theory}
and multi-shot solving \cite{gekakasc17a} supported by the \clingodl system \cite{janhunen2017clingo}.
The goal of the decomposition is to split highly complex scheduling tasks into
balanced portions for which partial schedules of good quality can be reliably
found within tight runtime limits. Then, the partial schedules are merged into a global solution of significantly better quality than obtainable in similar runtime with single-shot optimization on the full problem.
We address computational efficiency as well as solution quality by devising and empirically assessing decomposition strategies regarding the size of
time windows and heuristics for selecting their operations.

\subsection{Related Work}

While decision versions of scheduling problems can be successfully modeled and
solved by means of ASP modulo DL,
implemented by \clingodl on top of the
(multi-shot) ASP system \clingo \cite{gekakasc17a},
the optimization capacities of \clingodl come to their limits on moderately sized
yet highly combinatorial JSP instances \cite{el2020job}, for some of which optimal solutions are so far unknown \cite{shysha18a}.
Successful applications in areas beyond JSP include, e.g.,
industrial printing \cite{balduccini11a},
team-building \cite{rigralmaliiile12a},
matchmaking \cite{geglsasc13a},
shift design \cite{abseher2016shift},
course timetabling \cite{bainkaokscsotawa18a},
workload smoothing \cite{sabsim20a},
and
medical treatment planning \cite{dogagrmamopo21a},
pointing out the general attractiveness of ASP for modeling and solving scheduling problems.

In real-world production scheduling, the number of operations to process can easily go
into tens of thousands \cite{da2022industrial,kohakamo20a,kotakoscge21a}, which exceeds exact
optimization capacities even of state-of-the-art solvers for ASP,
Mixed Integer Programming (MIP), or Constraint Programming (CP)
\cite{daneshamooz2021mathematical,francescutto2021solving,shi2021solving}.
Hence, more efficient approaches to approximate good-quality schedules 
instead of striving for optimal solutions have attracted broad research interest.
On the one hand, respective methods include greedy and local search techniques such as
dispatching rules \cite{blackstone1982state}, shifting bottleneck \cite{adams1988shifting} and
genetic algorithms \cite{pezzella2008genetic}.
On the other hand, problem decomposition strategies based on a
rolling horizon \cite{singer2001decomposition,liu2008prediction} or
bottleneck operations \cite{zhang2010hybrid,zhai2014decomposition} 
have been proposed to partition large-scale instances into better manageable subproblems,
where no single strategy strictly dominates in minimizing the tardiness w.r.t.\ deadlines \cite{ovacik2012decomposition}.

\subsection{Main Contributions}

The contributions of our work going beyond the studies in \cite{el2022problem,elscge22a} are:

\begin{itemize}
	\item We develop a comprehensive multi-shot ASP modulo DL
	framework for JSP solving and significantly extend the
	preliminary presentation in~\cite{el2022problem}.
	In addition to a decomposition strategy based on the
	earliest starting times of operations, we describe how their
	remaining processing times can be used instead and further
	detail the ASP encoding of two-layered decomposition approaches,
	combining either basic strategy with the consideration of bottleneck machines.
	
	\item Since a decomposition into time windows may be incompatible with
	the optimal sequences of operations sharing a machine, we incorporate
	\textit{overlapping} time windows into the iterative scheduling process to
	offer chances for revising ``decomposition mistakes''.
	Moreover, the makespan objective, which we apply to optimize (partial) schedules, tolerates unnecessary idle times of machines as long as they do not yield a greater scheduling horizon. 
	Hence, we 
	model a \textit{compression} strategy in ASP to postprocess partial schedules by reassigning operations to earlier idle slots available on their machines.
	The ASP programs providing declarative implementations of
	the overlapping and compression techniques have not been
	elaborated before.
	
	\item Considering that static properties of JSP instances, describing the jobs, their operations, and available machines,
	merely provide rough information for guiding problem decomposition by basic features
	\cite{el2022problem}
	or clustering methods \cite{elscge22a},
	we further explore greedy search methods given by the \textit{First-In-First-Out}, \textit{Most-Total-Work-Remaining}, and \textit{Reinforcement Learning} dispatching heuristics proposed in \cite{tassel2021reinforcement} to assign operations to time windows based on the greedy solutions.
	
	\item We experimentally evaluate decomposition strategies varying the number of
	operations per time window as well as the
	overlapping and compression techniques
	to apply during multi-shot ASP modulo DL solving
	on Taillard's and Demirkol's well-known JSP benchmark sets 
	\cite{taillard1993benchmarks,demirkol1998benchmarks}.
	Our experiments show that the successive optimization of time windows
	leads to substantially better schedules within
	tight runtime limits than single-shot ASP modulo DL optimization on the full problem.
	In particular, taking greedy solutions obtained with the aforementioned dispatching heuristics 
	as basis for problem decomposition yields improved solution quality.
	\addtext{%
    To assess the scalability of our multi-shot ASP modulo DL methods,
	we extend the scope to a benchmark set of industrial-size JSP instances due to Da Col and Teppan~\cite{da2022industrial},
	and further contrast the obtained results with the state of the art established by high-performance CP systems.}
	
\end{itemize}
\noindent 
The paper is organized as follows.
Section~\ref{sec:preliminaries} briefly introduces ASP along with the relevant
extensions of multi-shot solving and DL constraints. In Section~\ref{sec:problem}, we present our successive optimization approach,
incorporating ASP programs encoding problem decomposition or iterative scheduling based on time windows, respectively,
as well as a decomposition scheme by constrained clustering.
Section~\ref{sec:experiments} provides experimental results on JSP benchmark sets, assessing different decomposition strategies along with the impact of overlapping and compression techniques.
Conclusions and future work are discussed in Section~\ref{sec:conclusions}.  

\section{Preliminaries}\label{sec:preliminaries}

Answer Set Programming (ASP) \cite{lifschitz19a} is a knowledge representation and reasoning paradigm geared for the effective 
modeling and solving of combinatorial (optimization) problems. A (first-order) ASP program consists of \emph{rules} of the form \lstinline{h :- b}$_1$\lstinline{,}$\dots$\lstinline{,b}$_n$\lstinline{.},
in which the head \lstinline{h} is an atom
\lstinline{p(t}$_1$\lstinline{,}$\dots$\lstinline{,t}$_m$\lstinline{)} or a
choice
\lstinline{{p(t}$_1$\lstinline{,}$\dots$\lstinline{,t}$_m$\lstinline|)}|
and each body literal \lstinline{b}$_i$ is an atom
\lstinline{p(t}$_1$\lstinline{,}$\dots$\lstinline{,t}$_m$\lstinline{)},
possibly preceded by the default negation connective \lstinline{not} and/or
followed by a condition
\lstinline|: c|$_1$\lstinline{,}$\dots$\lstinline{,c}$_l$, 
a built-in comparison \lstinline{t}$_1\circ{}$\lstinline{t}$_2$
with $\circ\in\{\text{\lstinline{<}},\text{\lstinline{<=}},\text{\lstinline{=}},\text{\lstinline{!=}},\text{\lstinline{>=}},\text{\lstinline{>}}\}$,
or an aggregate
\lstinline|t|$_0$\lstinline| = #count{t|$_1$\lstinline{,}$\dots$\lstinline|,|\linebreak[1]\lstinline|t|$_m$%
	\lstinline| : c|$_1$\lstinline{,}$\dots$\lstinline{,c}$_l$\lstinline|}|.
Each \lstinline|t|$_j$ denotes a term, i.e., a constant, variable, tuple,
or arithmetic expression, and each element \lstinline|c|$_k$ of a condition is
an atom that may be preceded by \lstinline{not} or a built-in comparison.
Roughly speaking, an ASP program is a shorthand for its ground instantiation,
obtainable by substituting variables with all of the available constants
and evaluating arithmetic expressions,
and the semantics is given by \emph{answer sets}, i.e.,
sets of (true) ground atoms such that all rules of the ground instantiation
are satisfied and allow for deriving each of the ground atoms 
by 
some rule whose body is satisfied.
The syntax of the considered ASP programs is a fragment of the modeling languages
described in \cite{cafageiakakrlemarisc20a,gehakalisc15a},
the ground instantiation process is detailed in \cite{kamsch21a},
and the answer set semantics is further elaborated in \cite{gehakalisc15a,lifschitz19a}.

For example, the ASP program
\begin{lstlisting}[numbers=none,frame=none]
{p(1..2)}.
p(0) :- not p(X) : X = 1..2.
p(X) :- X = #count{Y : p(Y), Y < 2}.
\end{lstlisting}
with the variables \lstinline{X} and \lstinline{Y},
the arithmetic expression \lstinline{1..2} standing
for the integer interval $[1,2]$, and the
built-in comparisons \lstinline{X = 1..2} and \lstinline{Y < 2}
is a shorthand for the following ground instantiation:%
\begin{lstlisting}[numbers=none,frame=none]
{p(1)}.
{p(2)}.
p(0) :- not p(1) : 1 = 1; not p(2) : 2 = 2.
p(0) :- 0 = #count{0 : p(0), 0 < 2; 
                   1 : p(1), 1 < 2;
                   2 : p(2), 2 < 2}.
p(1) :- 1 = #count{0 : p(0), 0 < 2; 
                   1 : p(1), 1 < 2; 
                   2 : p(2), 2 < 2}.
p(2) :- 2 = #count{0 : p(0), 0 < 2;
                   1 : p(1), 1 < 2;
                   2 : p(2), 2 < 2}.
\end{lstlisting}
\addtext{%
The first two rules with the choices
\lstinline{\{p(1)\}} and \lstinline{\{p(2)\}}
express that each of the atoms \lstinline{p(1)} and \lstinline{p(2)} may,
but does not necessarily have to be true.
As the built-in comparisons in the conditions
\lstinline{: 1 = 1} and \lstinline{: 2 = 2} of the third rule hold,
this rule derives the atom \lstinline{p(0)} when both
\lstinline{p(1)} and \lstinline{p(2)} are false.
In that case, the aggregate in the body of the fifth rule}
\begin{lstlisting}[numbers=none,frame=none,basicstyle=\ttfamily\small\redcolor]
p(1) :- 1 = #count{0 : p(0), 0 < 2; 
                   1 : p(1), 1 < 2; 
                   2 : p(2), 2 < 2}.
\end{lstlisting}
\addtext{holds, while the head atom \lstinline{p(1)} is false,
so that the rule is not satisfied.
This in turn means that at least one of the atoms
\lstinline{p(1)} and \lstinline{p(2)} needs to be true,
where having \lstinline{p(2)} alone contradicts with the fourth rule:}
\begin{lstlisting}[numbers=none,frame=none,basicstyle=\ttfamily\small\redcolor]
p(0) :- 0 = #count{0 : p(0), 0 < 2; 
                   1 : p(1), 1 < 2;
                   2 : p(2), 2 < 2}.
\end{lstlisting}
\addtext{%
If \lstinline{p(0)} is false, the aggregate in the body holds and
the rule is not satisfied. On the other hand, the condition \lstinline{: p(0), 0 < 2}
applies in case \lstinline{p(0)} is true, so that the rule body is not satisfied and \lstinline{p(0)}
turns out to be underivable.
Unlike that, the aggregate does not hold when \lstinline{p(1)} is true,
which makes the atom \lstinline{p(0)} underivable and \lstinline{p(2)} optional,
considering that the satisfaction of the rules}
\begin{lstlisting}[numbers=none,frame=none,basicstyle=\ttfamily\small]
p(1) :- 1 = #count{0 : p(0), 0 < 2; 
                   1 : p(1), 1 < 2; 
                   2 : p(2), 2 < 2}.
p(2) :- 2 = #count{0 : p(0), 0 < 2;
                   1 : p(1), 1 < 2;
                   2 : p(2), 2 < 2}.
\end{lstlisting}
\addtext{%
is readily established once \lstinline{p(0)} is false.
Hence, we obtain the two answer sets $\{\text{\lstinline{p(1)}}\}$ and
$\{\text{\lstinline{p(1)}},\text{\lstinline{p(2)}}\}$,
which satisfy all ground rules, where \lstinline{p(1)} and
\lstinline{p(2)} are also derivable in view of the choice rules
\lstinline{\{p(1)\}.} and \lstinline{\{p(2)\}.}}

Multi-shot ASP solving \cite{gekakasc17a}
allows for iterative reasoning processes by controlling and interleaving
the grounding and search phases of a stateful ASP system.
For referring to a collection of rules to instantiate,
the input language of \clingo supports \lstinline{#program name(c).}
directives, where \lstinline{name} denotes a \emph{subprogram} comprising
the rules below such a directive and the parameter~\lstinline{c} is a
placeholder for some value, e.g., the current time step in case of a
planning problem, supplied upon instantiating the subprogram.
Moreover, \lstinline{#external h : b}$_1$\lstinline{,}$\dots$\lstinline{,b}$_n$\lstinline{.}
statements are formed similar to rules yet declare an atom~\lstinline{h} as
\emph{external} when the body is satisfied:
such an external atom can be freely set to true or false by means of the
Python interface of \clingo,
so that rules including it in the body can be selectively (de)activated
to direct the search.\\
\noindent 
ASP modulo DL integrates DL constraints \cite{cotmal06a}, i.e., expressions written as
\lstinline|&diff{t|$_1$\lstinline| - t|$_2$\lstinline|}| \lstinline|<= t|$_3$, in the head of rules.
With the exception of the constant~\lstinline{0}, which denotes the number zero,
the terms \lstinline|t|$_1$ and \lstinline|t|$_2$ represent DL variables that can be
assigned any integer value.
However, the difference \lstinline|t|$_1$\lstinline| - t|$_2$ must not exceed the
integer constant~\lstinline|t|$_3$ if the body of a rule with the DL constraint in the
head is satisfied.
That is, the DL constraints asserted by rules whose body is satisfied restrict the
feasible values for DL variables,
and the \clingodl system extends \clingo by assuring the consistency of DL constraints
imposed by an answer set.
If these DL constraints are satisfiable, a canonical assignment of smallest feasible
integer values to DL variables can be determined in polynomial time and is output
together with the answer set.

\section{Multi-shot JSP Solving}
\label{sec:problem}

This section describes our successive optimization approach to
JSP solving by means of multi-shot ASP with \clingodl.
We start with specifying the fact format for JSP instances,
then detail several problem decomposition strategies,
including simple approaches based on static properties like the earliest starting times of operations as well as more elaborate
feature extraction and constrained clustering schemes,
present our ASP encoding with DL constraints for optimizing the makespan of partial schedules, and
finally outline the iterative scheduling process 
along with the incorporation of time window overlapping and compression techniques.

\subsection{Problem Instance}\label{subsec:instance}
Each job in a JSP instance is a sequence of operations with associated
machines and processing times.
Corresponding facts for an example instance with three jobs and three
machines are displayed in Listing~\ref{prg:facts}.
%
\begin{lstlisting}[float=t,label=prg:facts,caption={Example JSP instance},linerange={1-3},numbers=none,xleftmargin=3pt]
operation(1,1,1,3).  operation(2,1,2,4).  operation(3,1,3,9). #(\label{prg:facts:ops:begin}#)
operation(1,2,2,3).  operation(2,2,1,6).  operation(3,2,1,3).
operation(1,3,3,1).  operation(2,3,3,2).  operation(3,3,2,8). #(\label{prg:facts:ops:end}#)
\end{lstlisting}
An atom of the form
\lstinline{operation(}$j$\lstinline{,}$s$\lstinline{,}$m$\lstinline{,}$p$\lstinline{)}
denotes that the step~$s$ of job~$j$ needs to be processed by machine~$m$ for $p$ time units.
For example, the second operation of job~\lstinline{3} has a processing time of
\lstinline{3} time units on machine~\lstinline{1},
as specified by the fact
\lstinline{operation(3,}\linebreak[1]\lstinline{2,}\linebreak[1]\lstinline{1,3).}
The operation cannot be performed before the first operation of job~\lstinline{3}
is completed,
and its execution must not intersect with the first operation of job~\lstinline{1}
or the second operation of job~\lstinline{2},
which need to be processed by machine~\lstinline{1} as well.
That is, a schedule for the example instance must determine a sequence in which
to process the three mentioned operations on machine~\lstinline{1},
and likewise for operations sharing machine~\lstinline{2} or~\lstinline{3}, respectively.

\begin{figure}[t]
	\begin{tikzpicture}[x=3.5ex,y=3.5ex,label distance=-1ex,thick]
		\SetScales
		\foreach \i in {1,...,3} {
			\node[label=left:Machine \i] at (0,3.5-\i) {};
		}
		\foreach \i in {0,...,20} {
			\draw[dotted] (\i,0) -- (\i,3) node [below] at (\i,0) {$\i$};
		}
		\foreach \i in {0,...,3} {
			\draw[dotted] (0,\i) -- (20,\i);
		}
		\node[minimum width=3\scaledx-0.05\scaledx,minimum height=1\scaledy-0.05\scaledy,inner sep=0,rectangle,draw,anchor=south west,pattern color=lightgray,pattern=north west lines] at (0,2) {\textbf{1}-\textbf{1}};
		\node[minimum width=3\scaledx-0.05\scaledx,minimum height=1\scaledy-0.05\scaledy,inner sep=0,rectangle,draw,anchor=south west,pattern color=lightgray,pattern=crosshatch] at (9,2) {\textbf{3}-\textbf{2}};
		\node[minimum width=6\scaledx-0.05\scaledx,minimum height=1\scaledy-0.05\scaledy,inner sep=0,rectangle,draw,anchor=south west,pattern color=lightgray,pattern=grid] at (12,2) {\textbf{2}-\textbf{2}};
		\node[minimum width=4\scaledx-0.05\scaledx,minimum height=1\scaledy-0.05\scaledy,inner sep=0,rectangle,draw,anchor=south west,pattern color=lightgray,pattern=grid] at (0,1) {\textbf{2}-\textbf{1}};
		\node[minimum width=3\scaledx-0.05\scaledx,minimum height=1\scaledy-0.05\scaledy,inner sep=0,rectangle,draw,anchor=south west,pattern color=lightgray,pattern=north west lines] at (4,1) {\textbf{1}-\textbf{2}};
		\node[minimum width=8\scaledx-0.05\scaledx,minimum height=1\scaledy-0.05\scaledy,inner sep=0,rectangle,draw,anchor=south west,pattern color=lightgray,pattern=crosshatch] at (12,1) {\textbf{3}-\textbf{3}};
		\node[minimum width=9\scaledx-0.05\scaledx,minimum height=1\scaledy-0.05\scaledy,inner sep=0,rectangle,draw,anchor=south west,pattern color=lightgray,pattern=crosshatch] at (0,0) {\textbf{3}-\textbf{1}};
		\node[minimum width=1\scaledx-0.05\scaledx,minimum height=1\scaledy-0.05\scaledy,inner sep=0,rectangle,draw,anchor=south west,pattern color=lightgray,pattern=north west lines] at (9,0) {\textbf{1}-\textbf{3}};
		\node[minimum width=2\scaledx-0.05\scaledx,minimum height=1\scaledy-0.05\scaledy,inner sep=0,rectangle,draw,anchor=south west,pattern color=lightgray,pattern=grid] at (18,0) {\textbf{2}-\textbf{3}};
	\end{tikzpicture}
	\figspace
	\caption{Optimal schedule for example JSP instance in Listing~\ref{prg:facts}\label{fig:schedule}}
\end{figure}

Figure~\ref{fig:schedule} depicts a schedule with the optimal makespan, i.e., the latest completion time of any job/operation, for the JSP instance from Listing~\ref{prg:facts}.
The \textbf{J}-\textbf{S} pairs in horizontal bars indicated for the
machines~\lstinline{1}, \lstinline{2}, and~\lstinline{3} identify operations
by their job~\textbf{J} and step number~\textbf{S}.
For each machine, observe that the bars for operations it processes do not
intersect, so that the operations are performed in sequential order.
Moreover, operations belonging to the same job are scheduled one after another.
For example,
the second operation of job~\lstinline{3} is started after the
completion of its predecessor operation at time~$9$,
regardless of the availability of machine~\lstinline{1}
from time~$3$ on.
As the precedence of operations within their jobs must be respected
and the sum of processing times for operations of job~\lstinline{3}
matches the makespan~$20$, it is impossible to reduce the scheduling
horizon any further,
which yields that the schedule shown in Figure~\ref{fig:schedule} is optimal.

\subsection{Problem Decomposition}\label{subsec:decomposition}
Since JSP instances are highly combinatorial and the ground representation size
can also become problematic for large real-world scheduling problems, achieving
scalability of complete optimization methods necessitates problem decomposition.
In order to enable a successive extension of good-quality partial schedules to a global solution,
we consider strategies for partitioning the operations of JSP instances
into balanced time windows, each comprising an equal number of operations such that their precedence
within jobs is respected. In the following, we first detail problem decomposition based on the earliest starting times of operations, and then outline further strategies that can be encoded by stratified ASP programs having a unique answer set \cite{przymusinski1988declarative} as well.

Our encoding for \emph{Job-based Earliest Starting Time (J-EST)} decomposition
in Listing~\ref{prg:decomposition} takes a JSP instance specified by facts over
\lstinline{operation/4} as input.
In addition, a constant~\lstinline{n},
set to the default value~\lstinline{2} in line~\ref{prg:predeco:tw:begin},
determines the number of time windows into which the given operations
shall be split.
As we aim at time windows of (roughly) similar size,
the target number of operations per time window is in line~\ref{prg:predeco:optw:begin}
calculated by $\lceil \text{\lstinline{N}} / \text{\lstinline{n}}\rceil$,
where \lstinline{N} is the total number of operations.
For example, we obtain \lstinline{width(5)} for partitioning the nine operations
of the JSP instance in Listing~\ref{prg:facts} into two time windows.

\begin{lstlisting}[float=t,label=prg:decomposition,caption={J-EST decomposition encoding},linerange={1-10},xleftmargin=1.5em,fontadjust =true, basewidth= {0.57em, 0.45em}]
#const n = 2.  #(\label{prg:predeco:tw:begin}#)
width((N + n - 1) / n) :- N = #count{J,S : operation(J,S,M,P)}. #(\label{prg:predeco:optw:begin}#)

est(J,1,P,0)      :- operation(J,1,M,P).        #(\label{prg:assignment_process:est:begin}#)
est(J,S,P,P' + T) :- operation(J,S,M,P), est(J,S - 1,P',T).  #(\label{prg:assignment_process:est:end}#)

index(J,S,N) :- est(J,S,P,T),  #(\label{prg:assignment_process:rank:begin}#)
       N = #count{J',S' : est(J',S',P',T'), (T',P',J') < (T,P,J)}.   #(\label{prg:assignment_process:rank:end}#)

window(J,S,(N + W) / W) :- index(J,S,N), width(W). #(\label{prg:assignment_process:assign:begin}#) #(\label{prg:assignment_process:assign:end}#)
\end{lstlisting}

The rules in lines~\ref{prg:assignment_process:est:begin} and~\ref{prg:assignment_process:est:end}
encode the J-EST calculation per operation of a job,
given by the sum of processing times for predecessor operations belonging to the same job.
This yields, e.g.,
\lstinline{est(3,1,9,0)},
\lstinline{est(3,2,3,9)}, and
\lstinline{est(3,3,8,12)}
for the three operations of job~\lstinline{3} in our example instance,
where the third argument of an atom over \lstinline{est/4} provides
the processing time and the fourth the earliest starting time of an operation.
Note that the obtained earliest starting times match the first feasible time
points for scheduling operations and do thus constitute an optimistic
estimation of when to process the operations.

With the earliest starting times of operations at hand,
the rule in lines \ref{prg:assignment_process:rank:begin}-\ref{prg:assignment_process:rank:end}
determines a \emph{total order} of operations in terms of consecutive indexes ranging
from~\lstinline{0}.
That is, each operation is mapped to the number of operations with
(i) a smaller earliest starting time,
(ii) the same earliest starting time and shorter processing time, or
(iii) a smaller job identifier as tie-breaker in case of identical
earliest starting and processing times.
For the example JSP instance in Listing~\ref{prg:facts},
we obtain the indexes~\lstinline{0} to~\lstinline{2} for the first operations of the
three jobs,
the indexes~\lstinline{3} and~\lstinline{4} for the second operation of
job~\lstinline{1} or~\lstinline{2}, respectively,
in view of their earliest starting times~\lstinline{3} and~\lstinline{4},
and indexes from~\lstinline{5} to~\lstinline{8} for the remaining operations.
A relevant condition that is guaranteed by such a total order is that
indexes increase according to the precedence of operations within their jobs,
given that the earliest starting times grow along the sequence of operations in a job.

The last rule in line~\ref{prg:assignment_process:assign:begin} inspects the total
order of operations to partition them into time windows of the size~\lstinline{W}
determined by \lstinline{width(W)},
where only the last time window may possibly include fewer operations in case the
split is uneven.
As the ASP program encoding problem decomposition is stratified, its ground
instantiation can be simplified to (derived) facts, as shown in Listing~\ref{prg:tw}
for our example instance.
Time window numbers from~\lstinline{1} to~\lstinline{n = 2} are given by
the third argument of atoms over \lstinline{window/3}, so that the second operation
of job~\lstinline{3} and the third operation of each job form the time window~\lstinline{2},
while time window~\lstinline{1} consists of the five other operations.
\begin{lstlisting}[float=b,label=prg:tw,caption={Example time windows},numbers=none,xleftmargin=3pt]
window(1,1,1).  window(2,1,1).  window(3,1,1).
window(1,2,1).  window(2,2,1).  window(3,2,2).
window(1,3,2).  window(2,3,2).  window(3,3,2).
\end{lstlisting}

In addition to J-EST decomposition, we have devised a similar ASP program for
\emph{Job-based Most Total Work Remaining (J-MTWR)} decomposition,
where the total order of operations is decreasing by the sum of processing
times for an operation and its successors in a job.
The J-MTWR strategy is encoded by replacing the rules in lines
\ref{prg:assignment_process:est:begin}-\ref{prg:assignment_process:rank:end} of Listing~\ref{prg:decomposition}
by:
\begin{lstlisting}[numbers=none,frame=none,breaklines=false]
rpt(J,S,P)     :- operation(J,S,M,P), not operation(J,S + 1,_,_).
rpt(J,S,P + T) :- operation(J,S,M,P), rpt(J,S + 1,T).

index(J,S,N) :- rpt(J,S,T),
	N = #count{J',S' : rpt(J',S',T'), (T,J') < (T',J)}.
\end{lstlisting}
For example, we obtain the J-MTWR values~\lstinline{7}, \lstinline{12}, and~\lstinline{20},
represented by the atoms
\lstinline{rpt(1,1,7)}, \lstinline{rpt(2,1,12)}, and
\lstinline{rpt(3,1,20)},
for the first operation of job~\lstinline{1}, \lstinline{2}, or~\lstinline{3},
respectively, for the JSP instance in Listing~\ref{prg:facts},
matching the time for executing all three operations of each job.
Hence, the first operation of job~\lstinline{3} is considered as the most
important and is associated with the index~\lstinline{0},
and one can check that all three operations of job~\lstinline{3}
together with the first two operations of job~\lstinline{2}
form the first of \lstinline{n = 2} time windows, so that the
obtained decomposition varies from the time windows of J-EST
in Listing~\ref{prg:tw}. 
However, as J-MTWR values are decreasing along the sequence of operations in a job, the resulting operation indexes and time windows also respect the precedence of operations.

Beyond partitioning operations in a purely Job-based fashion,
we have encoded \emph{Machine-based} decompositions \emph{M-EST} and \emph{M-MTRW} in which an operation from a bottleneck machine with the greatest sum of processing times for yet unordered operations is of highest priority.
To this end, the encoding in Listing~\ref{prg:machine} builds on the
atoms over \lstinline{index/3} as obtained with either the J-EST or
J-MTWR strategy, but then it determines a bottleneck machine from
which the next operation in the underlying Job-based priority is picked and inserted into the total order of operations to be used for decomposition into time windows.
As long as some operation is yet unordered,
the rules in lines \ref{prg:machine:todo1}-\ref{prg:machine:done2}
group the operations into those that still need to be inserted or not, indicated by atoms over \lstinline{todo/4} or \lstinline{done/4}, respectively, whose third argument provides the Job-based index of an operation and the fourth an iteration counter for determining
a bottleneck machine w.r.t.\ the processing times for still unordered operations.
For our example instance in Listing~\ref{prg:facts}, the sum of processing times for operations 
is~\lstinline{12}, \lstinline{15}, or again \lstinline{12}, respectively,
for the machines~\lstinline{1}, \lstinline{2}, and~\lstinline{3}.
%
\begin{lstlisting}[float=t,label=prg:machine,caption={Machine-based decomposition encoding},linerange={1-40},firstnumber=10, xleftmargin=1.5em]
todo(J,S,N,1)     :- index(J,S,N). #(\label{prg:machine:todo1}#)
todo(J,S,N,I + 1) :- todo(J,S,N,I), select(J',S',M,I), J' != J. #(\label{prg:machine:todo2}#)
todo(J,S,N,I + 1) :- todo(J,S,N,I), select(J,S',M,I), S' < S. #(\label{prg:machine:todo3}#)
todo(I)           :- todo(J,S,N,I). #(\label{prg:machine:todo4}#)

done(J,S,N,I) :- todo(I), insert(J,S,N,I - 1). #(\label{prg:machine:done1}#)
done(J,S,N,I) :- todo(I), done(J,S,N,I - 1). #(\label{prg:machine:done2}#)

load(M,J,P,I)     :- operation(J,S,M,P), todo(J,S,N,I), #(\label{prg:machine:load1a}#)
                     not operation(J + 1,_,_,_). #(\label{prg:machine:load1b}#)
load(M,J,0,I)     :- operation(J,S,M,P), done(J,S,N,I), #(\label{prg:machine:load2a}#)
                     not operation(J + 1,_,_,_). #(\label{prg:machine:load2b}#)
load(M,J,P + L,I) :- operation(J,S,M,P), todo(J,S,N,I), #(\label{prg:machine:load3a}#)
                     load(M,J + 1,L,I). #(\label{prg:machine:load3b}#)
load(M,J,L,I)     :- operation(J,S,M,P), done(J,S,N,I), #(\label{prg:machine:load4a}#)
                     load(M,J + 1,L,I). #(\label{prg:machine:load4b}#)

high(M,M,L,I)   :- load(M,1,L,I), not operation(_,_,M + 1,_). #(\label{prg:machine:high1}#)
high(M,M,L,I)   :- load(M,1,L,I), high(M + 1,M',L',I), L' <= L. #(\label{prg:machine:high2}#)
high(M,M',L',I) :- load(M,1,L,I), high(M + 1,M',L',I), L < L'. #(\label{prg:machine:high3}#)

screen(0,I)     :- todo(I). #(\label{prg:machine:screen1}#)
screen(N + 1,I) :- screen(N,I), done(J,S,N,I). #(\label{prg:machine:screen2}#)
screen(N + 1,I) :- screen(N,I), todo(J,S,N,I), #(\label{prg:machine:screen3a}#)
                   operation(J,S,M,P), high(1,M',L,I), M' != M. #(\label{prg:machine:screen3b}#)

select(J,S,N,I) :- screen(N,I), todo(J,S,N,I), #(\label{prg:machine:select1a}#)
                   operation(J,S,M,P), high(1,M,L,I). #(\label{prg:machine:select1b}#)
insert(J,S,N,I) :- select(J,S,N,I). #(\label{prg:machine:insert1}#)
insert(J,S,N,I) :- todo(J,S,N,I), insert(J,S + 1,N',I). #(\label{prg:machine:insert2}#)

next(0,1) :- todo(1). #(\label{prg:machine:next1}#)
next(R,I) :- todo(I), select(J,S,N,I - 1), result(J,S,R - 1). #(\label{prg:machine:next2}#)

result(J,1,R) :- insert(J,1,N,I), next(R,I). #(\label{prg:machine:result1}#)
result(J,S,R) :- insert(J,S,N,I), next(R,I), done(J,S - 1,N',I). #(\label{prg:machine:result2}#)
result(J,S,R) :- insert(J,S,N,I), result(J,S - 1,R - 1), #(\label{prg:machine:result3a}#)
                 insert(J,S - 1,N',I). #(\label{prg:machine:result3b}#)

window(J,S,(R + W) / W) :- result(J,S,R), width(W). #(\label{prg:machine:window}#)

#show window/3.
\end{lstlisting}

In the first iteration, where all operations are yet unordered,
the rules in lines \ref{prg:machine:load1a}-\ref{prg:machine:load4b}
that sum up the processing times per machine, yield the loads in
terms of the atoms
\lstinline{load(1,1,12,1)},
\lstinline{load(2,1,15,1)}, and
\lstinline{load(3,1,12,1)} with the respective machine as the
first argument.
Then, the rules in lines \ref{prg:machine:high1}-\ref{prg:machine:high3} traverse all machines to determine the one with the highest remaining load, which in the first iteration leads to the atoms 
\lstinline{high(3,3,12,1)},
\lstinline{high(2,2,15,1)}, and
\lstinline{high(1,2,15,1)} with the last one providing the outcome
that some operation processed by the bottleneck machine~\lstinline{2} is to
be inserted into the total order of operations next.
The rules in lines \ref{prg:machine:screen1}-\ref{prg:machine:screen3b} inspect operations in the order of their Job-based indexes, skipping over those that are already inserted or
not processed by the bottleneck machine of the current iteration.
This yields the atom \lstinline{screen(1,1)} along with
\lstinline{select(2,1,1,1)} by the rule in
lines \ref{prg:machine:select1a}-\ref{prg:machine:select1b},
expressing that the first operation
of job~\lstinline{2} with the J-EST/J-MTWR index~\lstinline{1}
(instead of the first operation of job~\lstinline{1} 
or~\lstinline{3}, respectively, which with index~\lstinline{0}
come first according to the J-EST and J-MTWR orderings)
is to be inserted in the first iteration.
In general, this may lead to the choice of an operation
such that some of its predecessor operations processed by other machines are yet unordered, 
and the rules in lines \ref{prg:machine:insert1}-\ref{prg:machine:insert2} indicate all of them by derived
atoms over \lstinline{insert/4}, while the first iteration
for our example yields \lstinline{insert(2,1,1,1)} only.
The final positions for operations to insert into the Machine-based order are then determined by the rules in lines
\ref{prg:machine:result1}-\ref{prg:machine:result3b},
which derive the atom \lstinline{result(2,1,0)} to assign the first
operation of job~\lstinline{2} the highest priority in the total order of operations.

Remaining operations are further associated with the positions
\lstinline{1} to \lstinline{8} in the third argument of atoms over
\lstinline{result/3} obtained in the following iterations,
and the corresponding decomposition into
time windows is as with Job-based decomposition strategies represented by atoms over \lstinline{window/3},
derived by the rule in line~\ref{prg:machine:window}.
That is, our stratified ASP program in Listing~\ref{prg:machine}
for (re)ordering operations based on bottleneck machines can be
transparently combined with any initial total order of operations given in terms of
atoms over \lstinline{index/3}, so that this encoding is sufficient
to switch from both the J-EST and J-MTWR decomposition strategies to M-EST or M-MTWR, respectively.

\subsection{Feature Extraction}\label{subsec:features}

While the static properties considered above constitute easily
applicable yet very rough decomposition criteria,
we now turn to more elaborate features providing the basis for
constrained clustering.
In fact,
the application of machine learning methods to scheduling problems requires a careful selection of data describing the hidden dependencies between operations of different jobs \cite{ismail2012production,nasiri2019data}.
Clustering methods, which we 
apply in our approach, are no exception to this. That is, a clustering method requires an informative set of features characterizing the jobs, their operations, and machines of a JSP instance to identify patterns resulting in a beneficial decomposition of the operations into time windows. 

Heuristic methods suggested in the literature \cite{koonce2000using,harrath2002genetic,shahzad2010discovering,ismail2012production,adibi2014clustering,nasiri2019data} characterize instances of scheduling problems based on the following features: priority, processing time, remaining processing time, machine load, and sequence position. 
Most of these approaches convert the quantitative feature values 
into qualitative attributes in order to obtain generic dispatching rules
that remain applicable to instances of different size, while we propose a clustering method that can be applied to feature values directly and does not require any problem-specific transformations. However, our clustering method for JSP instance decomposition requires all features to have numerical values, which permit the calculation of distance measures for estimating (dis)similarities between operations.

In detail, we consider the following features of jobs, operations, and machines:
\begin{description}
	\item[Operation (OP)] is the ordinal value for the position of an operation in its job.
	
	\item[Processing Time (PT)] is the time for executing an operation on its machine.
	
	\item[Remaining Processing Time (RPT)] provides the total processing time for pending operations until the completion of a job. For example, Table~\ref{tab1} lists \textit{RPT} values for operations of the example instance in Listing~\ref{prg:facts}. The job $1$ consists of $3$ operations with a total processing time of $7$ time units, which matches \textit{RPT} for its first operation $O_{1,1}$. The \textit{RPT} for $O_{1,2}$ is obtained by subtracting the processing time of $O_{1,1}$, i.e., $7-3=4$, and it corresponds to the processing time $1$ for the last operation $O_{1,3}$ of job~$1$.
	
	\item[Time Length of a Job (TLJ)] is the total processing time for operations of a job, which coincides with the \textit{RPT} value of the job's first operation and is more coarse-grained than the operation-specific \textit{RPT} feature.
	
	\item[Earliest Starting Time (EST)] represents the earliest possible time for executing an operation, given by the total processing time for the predecessor operations in its job.
	For the first operation of each job, the \textit{EST} value defaults to~$0$.
	
	\item[Machine Load (ML)] is a property describing how much time it takes to execute the operations assigned to a machine. Initially, \textit{ML} corresponds to the total processing time for all operations to be executed by a machine.
	Then the assumption is that the operations are processed in increasing order of their \textit{EST} values, and \textit{ML} is thus reduced by the processing times of preceeding operations.
	For example, the \textit{EST} for the operations $O_{3,1}$, $O_{1,3}$, and $O_{2,3}$ assigned to machine~$3$ is $0$, $6$, or $10$, respectively.
	Proceeding in this execution order, \textit{ML} is the total processing time $12$ for $O_{3,1}$, reduced by the processing time of $O_{3,1}$ to $12-9 = 3$ for $O_{1,3}$, and then we obtain the processing time $2$ for the last operation $O_{2,3}$.
	
	\item[Starting Time (ST)] is a family of features providing the starting times of operations obtained by scheduling them with heuristic search methods \addtext{based on the greedy algorithm shown in Figure~\ref{fig:greedy}.
	Its idea is to maintain a set $R$ of pairs $(O_{j,s},r)$ such that an operation $O_{j,s}$ represents the step $s$ of a job $j$ to be processed next and $r$ stands for the release time of $O_{j,s}$, i.e., the predecessor operation $O_{j,s-1}$ (if any) is completed and the machine of $O_{j,s}$ is available.
	Greedy allocation then proceeds by these release times,
	where the smallest time is denoted by~$t$ in step 1 of Figure~\ref{fig:greedy}.
	The crucial component of the algorithm is the selection heuristic
	in step~2, picking the next operation $O_{j,s}$ to allocate at time~$t$.}

	In our work, we consider Earliest Starting Time (\stest), First-In-First-Out (\stfifo), and Most Total Work Remaining (\stmtwr) as heuristics for the greedy operation allocation; see \cite{jones1998survey} for an overview of such techniques.
	In the case of \stfifo, the algorithm selects an operation waiting longest for its machine to become available.
	For example, as indicated in Table~\ref{tab1},
	the first operations $O_{1,1}$, $O_{2,1}$, and $O_{3,1}$ are allocated at time~$0$ (on different machines),
	then $O_{1,2}$ waits $1$ time unit for its machine~$2$ to get available and starts at time~$4$, and $O_{1,3}$ waits from time~$7$ for machine~$3$ and is started at time~$9$.
	While $O_{2,2}$ is started on machine~$1$ at time~$4$, so that
	$O_{2,3}$ is started on machine~$3$ at time~$10$,
	\addtext{$O_{3,2}$ needs to wait $1$ time unit to start on its machine~$1$ at time~$10$, and then $O_{3,3}$ is started on machine~$2$ at time~$13$.
	If the processing time of $O_{3,1}$ were $10$ instead of $9$,
	\stfifo{} would have the choice between allocating either $O_{1,3}$ or $O_{2,3}$
	on machine~$3$ at time~$10$.
	In view of the longer waiting time of $O_{1,3}$, it gets selected first and $O_{2,3}$ comes after its completion, which is how
	\stfifo{} aims at a fair allocation.}
	With the \stest and \stmtwr heuristics,
	the greedy algorithm's selection of the next operation to allocate
	is based on smaller \textit{EST} or
	greater \textit{RPT} values, respectively.
	\begin{figure}[t]
		\begin{tcolorbox}[arc=0mm,outer arc=0mm,colbacktitle=lightgray!50!white,coltitle=black,colframe=gray,fonttitle=\bfseries,colback=white,subtitle style={colback=white},left=1.3ex]
	\addarea{%
			Starting with $R:=\{(O_{j,1},0) \mid j \text{ is a job with first step } 1\}$,
			$S:=\emptyset$, and $T:=\emptyset$, \\ 
			while $R\neq\emptyset$:
			\begin{enumerate}[label=\textbf{\arabic*.}]
				\item $t:=\min(\{r \mid (O_{j,s},r) \in R\})$
				\item SELECT $(O_{j,s},t)\in R$ and set $R:=R\setminus\{(O_{j,s},t)\}$
				\item 
					If NOOP, 
					then 
					$S:=S\cup\{O_{j,s}\}$;
					else: 
					\begin{enumerate}[label=\textbf{\alph*.}]
						\item $T:=T\cup\{(O_{j,s},t)\}$
						\item $u:=t+p$ for the processing time $p$ of $O_{j,s}$ on machine $m$ 
						\item For each $(O_{j',s'},r)\in R$ such that $O_{j',s'}$ is processed by $m$, set \\ 
						$R:=(R\setminus\{(O_{j',s'},r)\}) \cup
									  \{(O_{j',s'},\max(\{r,u\}))\}$
						\item For each $O_{j',s'}\in S$ such that $O_{j',s'}$ is processed by $m$, set \\ 
						$R:=R\cup\{(O_{j',s'},u)\}$ and  $S:=S\setminus\{O_{j',s'}\}$
						\item If there is a step $s+1$ of job $j$ to be processed by machine $m'$, then \\
						$\begin{array}[t]{@{}l@{}l@{}}
									R:=R\cup{}&\{(O_{j,s+1},
											   \max(\{u\}
											   \cup {} \\ &
											   \{t'+p' \mid (O_{j',s'},t')\in T \text{ with processing time }p' \text{ on machine } m'\}))\}
						 \end{array}
						$
					\end{enumerate}
			\end{enumerate}
			If $S=\emptyset$, then return $T$}
		\end{tcolorbox}
		\caption{\addtext{Basic greedy operation allocation algorithm for computing a heuristic schedule}\label{fig:greedy}}
	\end{figure}
	\begin{table}[b]
		\setlength{\tabcolsep}{13.6pt}
		\centering
		\caption{Part of the features extracted for the example JSP instance in Listing~\ref{prg:facts}}
		\label{tab1}
		\begin{tabular}{r  r  r  r  r  r}
			\toprule
			\textbf{Operation} & \textbf{RPT} & \textbf{EST} & \textbf{ML} & \textbf{\stfifo} & \textbf{\wtfifo}\\
			\midrule
			$O_{1,1}$  & $7$ & $0$   & $12$	&  $0$  & $0$\\
			$O_{1,2}$  & $4$  & $3$   & $11$		&  $4$  & $1$\\
			$O_{1,3}$  & $1$  & $6$  & $3$	  &  $9$ & $2$
			\\[1.5mm]
			$O_{2,1}$  & $12$ & $0$   & $15$	&  $0$  & $0$\\
			$O_{2,2}$  & $8$  & $4$   & $9$	&  $4$  & $0$\\
			$O_{2,3}$  & $2$  & $10$  & $2$	  &  $10$ & $0$
			\\[1.5mm]
			$O_{3,1}$  & $20$ & $0$  & $12$	  &  $0$  & $0$\\
			$O_{3,2}$  & $11$  & $9$  & $3$		&  $10$  & $1$\\
			$O_{3,3}$  & $8$  & $12$  & $8$		&  $13$ & $0$\\
			\bottomrule
		\end{tabular}
	\end{table}

	\addtext{%
	The implementation of greedy heuristics \cite{tassel2021reinforcement}
	we utilize never performs a noop in step~3 of Figure~\ref{fig:greedy} for
	\stest, \stfifo, or \stmtwr, yet includes this option for dispatching by
	Reinforcement Learning, which we compare in Section~\ref{sec:experiments}.
	When allocating a selected operation $O_{j,s}$, it is added to the heuristic schedule~$T$ in step 3a,
	and its completion time~$u$, calculated in step 3b,
	is used to update the release times of operations processed by the same machine as $O_{j,s}$ in step~3c.
	If there is a successor operation $O_{j,s+1}$, step 3e takes it to replace $O_{j,s}$ in $R$, where the release time depends on the completion of $O_{j,s}$ at time~$u$ and machine availability.
	Without noops, the final heuristic schedule $T$,
	such as the one for \stfifo{} in Table~\ref{tab1}, is always complete,
	while skipped operations are otherwise reconsidered in step~3d after
	some allocation on their machine.}
	\item[Waiting Time (WT)] is also a family of features, where variants denoted by \wtest, \wtfifo, and \wtmtwr rely on schedules obtained with the corresponding ST heuristic, i.e., \stest, \stfifo, or \stmtwr.
	Given a schedule computed by the greedy algorithm,
	the waiting time of an operation is determined by the difference
	between its starting time and the time of completing the predecessor
	operation, or simply the starting time for the first operation of each job.
	For instance, the starting times with \stfifo listed in Table~\ref{tab1}
	yield the waiting times given in the \wtfifo column.
	In fact, $O_{1,2}$ and $O_{1,3}$ wait for $1$ or $2$ time units,
	respectively, for their machines~$2$ and $3$ to get available, and $O_{3,2}$ also needs to wait $1$ time unit for the completion of $O_{2,2}$ before its execution by machine~$1$.
\end{description}
We extract all of the features described above from a given JSP instance
and can thus use them as inputs to our decomposition method by means of constrained clustering presented in the following.

\subsection{Constrained Clustering}
\label{subsec:method}

Clustering algorithms are unsupervised learning methods whose goal is to partition a set of data objects into (disjoint) clusters such that each cluster gathers objects of high similarity, i.e.,
operations belonging to the same time window in our case. 
Such similarity is determined by some measure, e.g., Euclidean distance, based on features of each object, like the features of operations described in the previous section. 
However, the direct application of common clustering algorithms, such as K-Means~\cite{Forgy1965ClusterAO}, to scheduling problems is impractical since the partitioning does not take the sequence of operations in a job into account. For instance, a clustering algorithm may put $O_{1,1}$ and $O_{1,3}$ into the same and $O_{1,2}$ into another cluster, in which case the sequence of time windows becomes inconsistent and no compatible schedule exists.

We thus propose a constrained clustering algorithm that preserves sequences of operations by considering their order in the assignment to clusters. That is, the predecessors of an operation to be put into the $n$th cluster must be assigned within the clusters $1,\dots,n$. Also considering that our approach involves cluster-wise combinatorial optimization, the generation of large clusters risks to deteriorate the solving performance significantly.
In the extreme case, all operations could be put into a single cluster representing the entire problem instance. Hence, in addition to the similarity of operations, our decomposition method also aims at balancing the number of operations per cluster. 

\begin{algorithm}[t]
\caption{Constrained Clustering Algorithm}\label{alg1}
\begin{algorithmic}
	\Require $\mathit{operations}$, $\mathit{num\_clusters}$ 
	\State $\mathit{cluster\_capacity} \gets \ceil*{\frac{|\mathit{operations}|}{\mathit{num\_clusters}}}$
	\State Generate $\mathit{num\_clusters}$ many centroids
	\For{$n=1$ \To $\mathit{num\_clusters}$}
	\State $\mathit{clusters}[n] \gets \emptyset$
	\State $\mathit{current\_capacity} \gets \mathit{cluster\_capacity}$
	\While{$0 < \mathit{current\_capacity}$}
	\State Calculate distance between data objects and $n$th centroid   \Comment{Using Euclidean distance}
	\State $O_{i,j} \gets \text{Nearest data object from }\mathit{operations}$ 
	\Repeat
	\State $\mathit{current\_capacity} \gets \mathit{current\_capacity} - 1$
	\State $\mathit{operations} \gets \mathit{operations} \setminus \{O_{i,j}\}$
	\State $\mathit{clusters}[n] \gets \mathit{clusters}[n] \cup \{O_{i,j}\}$ \Comment{Assigning operation $O_{i,j}$ to $n$th TW}
	\State $j \gets j-1$
	\Until{$O_{i,j}\notin \mathit{operations}$} \Comment{Satisfying the precedence constraint}
	
	\State Update the $n$th centroid 
	\EndWhile
	\EndFor
\end{algorithmic}
\end{algorithm}
Algorithm~\ref{alg1} provides a pseudocode description of our constrained clustering algorithm.
Similar to K-Means, we assume that the algorithm gets the target number of clusters 
into which the operations shall be partitioned as input.
The cluster capacity, used for balancing the operations per cluster, is then obtained by dividing the total number of operations by the number of clusters.
Moreover, the clustering algorithm takes care of generating one initial centroid per cluster, given by randomly selected operations that are compatible with the precedence relation.
For example, when each job consists of $15$ operations and the target number of clusters is $3$,
the first centroid will be an operation at the first to fifth place of its job,
the second an operation from place six to ten, and the third an operation at the eleventh
or later place.
%
In order to populate each cluster, the algorithm inspects features to determine the Euclidean distance of each yet unassigned operation to the centroid of the current cluster and assigns the nearest operation to the cluster. To also preserve the precedence between operations, we additionally include any yet unassigned predecessor operations in the current cluster, and then update its centroid with the features of newly assigned operations. Whenever the cluster capacity is reached, the algorithm proceeds to the next cluster, and this decomposition process continues until all operations are assigned to clusters.

In order to identify the most promising features for distance calculation among those introduced in the previous subsection,
we suggest the following forward selection principle:
start with a small set of features, perform decomposition by Algorithm~\ref{alg1} with several seeds to generate the initial
centroids,
and iteratively solve JSP instances with the obtained time windows.
Then, evaluate the possible extensions by one more feature,
compare the quality of resulting schedules, and pick the best 
extended set of features.
This process continues until either 
(i) all features are selected 
or 
(ii) any extension by another feature leads to solutions of lower average quality.

\subsection{Problem Encoding}\label{subsec:encoding}

Given a JSP instance as in Listing~\ref{prg:facts} along with facts like those
in Listing~\ref{prg:tw} providing a decomposition into time windows,
the idea of successive schedule optimization is to consider time windows
one after the other and gradually extend a partial schedule that fixes the
operations of previous time windows.
In this process, we adopt the makespan as optimization objective for scheduling
the operations of each time window,
thus applying the rule of thumb that small scheduling horizons for
partial schedules are likely to lead towards a global solution with short makespan.
While we use DL variables to compactly represent the starting times of operations to schedule,
we assume that a partial schedule for the operations of previous time windows is reified
in terms of additional input facts of the form 
\lstinline{start((}$j$\lstinline{,}$s$\lstinline{),}$t$\lstinline{,}$w$\lstinline{).},
where $t$ is the starting time scheduled for the step~$s$ of job~$j$ at the previous
time window indicated by~$w$.%
\begin{lstlisting}[float=t,label=prg:encoding,caption={Multi-shot ASP modulo DL encoding},xleftmargin = 1.5em,fontadjust =true, basewidth= {0.59em, 0.45em}]
#program step(w).

use(M,w,w) :- operation(J,S,M,P), window(J,S,w). #(\label{prg:encoding:use1}#)
use(M,W,w) :- use(M,W,w - 1), not window(J,S,w) : 
              operation(J,S,M,P). #(\label{prg:encoding:use2}#)

share((J1,S1),(J2,S2),P1,P2,X,w) :- operation(J1,S1,M,P1), #(\label{prg:base:sharing_machine:begin}#)
                                    operation(J2,S2,M,P2),
                                    window(J1,S1,W),
                                    window(J2,S2,w), 
                                    (W,J1) < (w,J2),
                                    use(M,W,w - X), X = 0..1. #(\label{prg:base:sharing_machine:end}#)

order(O1,O2,P1,w)        :- share(O1,O2,P1,P2,1,w). #(\label{prg:base:ordered}#) 
order((J,S1),(J,S2),P,w) :- operation(J,S1,M,P), window(J,S2,w), #(\label{prg:base:same_job:begin}#) 
                            S1 = S2 - 1. #(\label{prg:base:same_job:end}#)

{order(O1,O2,P1,w)} :- share(O1,O2,P1,P2,0,w). #(\label{prg:base:order}#)
 order(O2,O1,P2,w)  :- share(O1,O2,P1,P2,0,w), #(\label{prg:base:order:begin}#)
                       not order(O1,O2,P1,w).  #(\label{prg:base:order:end}#)

&diff{O - 0} <=  T :- start(O,T,w - 1). #(\label{prg:diff_log:freez1:begin}#) #(\label{prg:diff_log:freez1:end}#)
&diff{0 - O} <= -T :- start(O,T,w - 1). #(\label{prg:diff_log:freez2:begin}#) #(\label{prg:diff_log:freez2:end}#)

&diff{0 - (J,1)} <=  0 :- window(J,1,w). #(\label{prg:diff_log:non_nega:begin}#) #(\label{prg:diff_log:non_nega:end}#)
&diff{O1 - O2}   <= -P :- order(O1,O2,P,w). #(\label{prg:diff_log:oper_const:begin}#)

&diff{(J,S) - makespan} <= -P :- operation(J,S,M,P),window(J,S,w), #(\label{prg:diff_log:oper_limit:begin}#)
                                 not window(J,S + 1,w). #(\label{prg:diff_log:oper_limit:end}#)

#program optimize(m). #(\label{prg:diff_log:oper_min}#)
#external horizon(m).                                           #(\label{prg:diff_log:oper_min:begin}#)

&diff{makespan - 0} <= m :- horizon(m).                     #(\label{prg:diff_log:oper_min:end}#)
\end{lstlisting}

The \lstinline{step(w)} subprogram until line~\ref{prg:diff_log:oper_limit:end} 
in Listing~\ref{prg:encoding} constitutes the central part of our multi-shot ASP modulo DL encoding, whose parameter~\lstinline{w} stands for consecutive integers from~\lstinline{1}
identifying time windows to schedule.
Auxiliary atoms of the form \lstinline{use(}$m$\lstinline{,}$w'$\lstinline{,}$w$\lstinline{)},
supplied by the rules in lines~\ref{prg:encoding:use1} and~\ref{prg:encoding:use2},
indicate the latest time window $1\leq w'\leq w$ including some operation that needs
to be processed by machine~$m$.
The next rule in lines \ref{prg:base:sharing_machine:begin}-\ref{prg:base:sharing_machine:end}
identifies pairs \lstinline{(}$j_1$\lstinline{,}$s_1$\lstinline{)} and
\lstinline{(}$j_2$\lstinline{,}$s_2$\lstinline{)} of operations sharing
the same machine~$m$,
where
\lstinline{(}$j_2$\lstinline{,}$s_2$\lstinline{)} belongs to the time window~$w$ and
\lstinline{(}$j_1$\lstinline{,}$s_1$\lstinline{)} is either 
(i)
contained in the latest time window $1\leq w'< w$ indicated by
\lstinline{use(}$m$\lstinline{,}$w'$\lstinline{,}$w-1$\lstinline{)} or
(ii)
also part of the time window~$w$, in which case $j_1 < j_2$ establishes
an asymmetric representation for the pair of operations in derived atoms
\lstinline{share((}$j_1$\lstinline{,}$s_1$\lstinline{),(}$j_2$\lstinline{,}$s_2$\lstinline{),}$p_1$\lstinline{,}$p_2$\lstinline{,}$x$\lstinline{,}$w$\lstinline{)}.
If the flag $x=\text{\lstinline{1}}$ signals that 
\lstinline{(}$j_1$\lstinline{,}$s_1$\lstinline{)}
belongs to a previous time window~$w'$,
the rule in line~\ref{prg:base:ordered} 
derives the atom
\lstinline{order((}$j_1$\lstinline{,}$s_1$\lstinline{),(}$j_2$\lstinline{,}$s_2$\lstinline{),}$p_1$\lstinline{,}$w$\lstinline{)}
to express that \lstinline{(}$j_1$\lstinline{,}$s_1$\lstinline{)} needs to be completed
before performing \lstinline{(}$j_2$\lstinline{,}$s_2$\lstinline{)}, i.e.,
the execution order must comply with the decomposition into time windows.
The rule in lines \ref{prg:base:same_job:begin}-\ref{prg:base:same_job:end} yields
a similar atom when \lstinline{(}$j_1$\lstinline{,}$s_1$\lstinline{)}
is the predecessor operation $s_1 = s_2-1$ of \lstinline{(}$j_2$\lstinline{,}$s_2$\lstinline{)}
in the same job $j_1=j_2$.
In contrast to the cases in which 
\lstinline{(}$j_1$\lstinline{,}$s_1$\lstinline{)} must be processed before
\lstinline{(}$j_2$\lstinline{,}$s_2$\lstinline{)},
the choice rule in line~\ref{prg:base:order}
allows for performing two operations sharing a machine in the lexicographic order
of their jobs if the operations belong to the same time window.
In case the atom representing execution in lexicographic order is not chosen,
the rule in lines \ref{prg:base:order:begin}-\ref{prg:base:order:end}
derives an atom expressing the inverse, given that the operations must not intersect and some sequence has to be determined.
Note that the rules in lines \ref{prg:base:order}-\ref{prg:base:order:end} allow for choosing an execution order between
two operations
\lstinline{(}$j_1$\lstinline{,}$s_1$\lstinline{)} and
\lstinline{(}$j_2$\lstinline{,}$s_2$\lstinline{)} only if they
belong to the same time window~$w$ but distinct jobs $j_1<j_2$,
while the rules in lines \ref{prg:base:ordered}-\ref{prg:base:same_job:end} handle pairs of operations whose
execution order is fixed.

Given that the rules up to line~\ref{prg:base:order:end} yield atoms of the form
\lstinline{order(}$o_1$\lstinline{,}$o_2$\lstinline{,}$p_1$\lstinline{,}$w$\lstinline{)},
expressing the hard requirement (or choice) to perform an operation~$o_1$
with processing time~$p_1$ before an operation~$o_2$
that belongs to time window~$w$,
the remaining rules assert corresponding DL constraints. 
To begin with,
the starting times of operations~$o$ from the previous time window $w-1$ (if any)
are in the lines~\ref{prg:diff_log:freez1:begin} and~\ref{prg:diff_log:freez2:begin}
fixed by restricting them from above or below, respectively,
to the value~$t$ in an (optimized) partial schedule for time window $w-1$, 
as supplied by reified facts \lstinline{start(}$o$\lstinline{,}$t$\lstinline{,}$w-1$\lstinline{).}
In line~\ref{prg:diff_log:non_nega:begin},
the lower bound~$0$ is asserted for the starting time of the first operation of some job
included in the time window~$w$ for which a partial schedule is to be determined next.
In addition, DL constraints reflecting the order of performing operations are imposed 
in line~\ref{prg:diff_log:oper_const:begin},
which concerns operations sharing a machine as well as successor operations within jobs.
Since such constraints trace the sequence of operations in a job, they establish the
earliest starting time, considered for problem decomposition in Section~\ref{subsec:decomposition},
as lower bound for scheduling an operation, and the execution order on the machine
processing the operation can increase its starting time further.
The last rule of the \lstinline{step(w)} subprogram in lines
\ref{prg:diff_log:oper_limit:begin}-\ref{prg:diff_log:oper_limit:end} asserts
that the value for the DL variable \lstinline{makespan} cannot be less than the
completion time of any operation of the time window~$w$.
As a consequence, the least feasible \lstinline{makespan} value provides the
scheduling horizon of a partial schedule for operations of time windows up to~$w$.

The task of optimizing the horizon of a (partial) schedule means choosing
an execution order of operations of the latest time window sharing some machine
such that the value for the \lstinline{makespan} variable is minimized.
In single-shot ASP modulo DL solving with \clingodl, such minimization can be accomplished via the command-line option \lstinline{--minimize-variable=makespan}.
This option, however, is implemented by means of a fixed control loop that
cannot be combined with (other) multi-shot solving methods.
For the successive optimization based on time windows, where the scheduling
horizon gradually increases, we thus require a dedicated treatment of DL constraints
limiting the value for \lstinline{makespan}.
To this end, the \lstinline{optimize(m)} subprogram below line~\ref{prg:diff_log:oper_min}
declares an external atom \lstinline{horizon(m)} for controlling whether a DL
constraint asserted in line~\ref{prg:diff_log:oper_min:end} is active and limits
the \lstinline{makespan} value to an integer supplied for the parameter~\lstinline{m}.

\subsection{Iterative Scheduling}\label{subsec:scheduling}

The main steps of our control loop for successive schedule optimization
by multi-shot ASP modulo DL solving, implemented by means of the Python interface of \clingodl,
are displayed in Figure~{\ref{fig:enum}} and further detailed in the following.
When launching the optimization process for a new time window, any instances of the
\lstinline{horizon(m)} atom introduced before are set to false in step~1 for making sure that
some (partial) schedule $X$ is feasible.
No such atoms have been introduced yet for the first time window $w=1$,
where the static (default) subprogram called \lstinline{base},
supplying a JSP instance along with its decomposition in terms of facts over \lstinline{window/3},
and the \lstinline{step(}$1$\lstinline{)} subprogram for operations of the first time window
are instantiated in steps~2 and~3.
\begin{figure}[t]
	\begin{tcolorbox}[arc=0mm,outer arc=0mm,colbacktitle=lightgray!50!white,coltitle=black,colframe=gray,fonttitle=\bfseries,colback=white,subtitle style={colback=white},left=1.3ex]
		Starting with $H:=\emptyset$, for $w:= 1$ to number $n$ of time windows:
		\begin{enumerate}[label=\textbf{\arabic*.}]
			\item For each $h\in H$, set \lstinline{horizon(}$h$\lstinline{)} to false
			\item 
				If $w= 1$, 
				then 
				$P:=\{\text{\lstinline{base}}\}$; \\
				else 
				$P := \{\text{\lstinline{start((}$j$\lstinline{,}$s$\lstinline{),}$t$\lstinline{,}$w-1$\lstinline{).}}
				\mid
				\{\text{\lstinline{window(}$j$\lstinline{,}$s$\lstinline{,}$w-1$\lstinline{)}},
				\text{\lstinline{(}$j$\lstinline{,}$s$\lstinline{)}}=t\}$\rlap{${}\subseteq X\}$}%
			\item Ground $P\cup\{\text{\lstinline{step(}$w$\lstinline{)}}\}$
			\item While there is some answer set $X$ and the time limit per time window is not reached:
			\begin{enumerate}[label=\textbf{\alph*.}]
				\item $h:=t-1$, where $($\lstinline{makespan}${}=t)\in X$
				\item If $h\notin H$, then $H := H\cup\{h\}$ and ground $\{$\lstinline{optimize(}$h$\lstinline{)}$\}$
				\item Set \lstinline{horizon(}$h$\lstinline{)} to true
			\end{enumerate}
		\end{enumerate}
	\end{tcolorbox}
	\caption{Control loop for successive schedule optimization by multi-shot ASP modulo DL solving\label{fig:enum}}
\end{figure}

Once some schedule $X$ with a horizon $h+1$ is found in step~4a,
the step~4b consists of instantiating
the subprogram \lstinline{optimize(}$h$\lstinline{)} 
on demand, i.e.,
in case $h$ has not been passed as a value for~\lstinline{m} before, and
the corresponding \lstinline{horizon(}$h$\lstinline{)} atom is set to true in step~4c for activating the DL
constraint limiting the admitted scheduling horizon to~$h$.
The step 4 of successively reducing the horizon~$h$ in order to find better partial schedules
stops when the imposed \lstinline{makespan} value turns out to be infeasible,
meaning that an optimal partial schedule has been found.
As already mentioned above, the introduced instances of the external \lstinline{horizon(m)} atom
are then set to false in step~1 on the next iteration, and the successive optimization proceeds by in steps~2 and~3 instantiating the
\lstinline{step(}$w+\nolinebreak1$\lstinline{)} subprogram for the next time window $w+1$ (if any) and
also supplying the determined starting times of operations from time window~$w$
by reified facts.
The described control loop for successively extending good-quality partial schedules to
a global solution 
can be configured with a time limit,
included as secondary stopping condition in step~4,
to restrict the optimization efforts per time window
and thus make sure that the iterative scheduling progresses.

For illustrating some phenomena going along with problem decomposition
and iterative scheduling, let us inspect the schedule in Figure~\ref{fig:decomposed}
that can be obtained with the decomposition into windows given in Listing~\ref{prg:tw}.
The separation between the two time windows is indicated by bold double lines marking
the completion of the latest operation of the first time window on each of the three machines.
Notably, the partial schedule for the first time window as well as its extension to the
second time window are optimal in terms of their respective makespan.
However, the obtained global solution has a makespan of~$21$ rather than just~$20$
as for the optimal schedule in Figure~\ref{fig:schedule}.
The reason is that the second operation of job~\lstinline{3} would need to be
scheduled before the second operation of job~\lstinline{2} on machine~\lstinline{1},
while the decomposition into time windows dictates the inverse order and necessitates
the later completion of job~\lstinline{3}.
Given the available buffer for scheduling operations with comparably short processing
times on machine~\lstinline{3},
the third operation of job~\lstinline{1} can be performed after the third operation
of job~\lstinline{2} without deteriorating the makespan,
yet introducing an unnecessary idle time from~$9$ to~$10$ on machine~\lstinline{3} 
that could be avoided by choosing the inverse execution order.
Even though this may seem negligible for the example instance at hand,
idle slots can potentially propagate when a partial schedule gets extended to later time windows.%
\begin{figure}[b]
	\begin{tikzpicture}[x=3.4ex,y=3.5ex,label distance=-1ex,thick]
		\SetScales
		\foreach \i in {1,...,3} {
			\node[label=left:Machine \i] at (0,3.5-\i) {};
		}
		\foreach \i in {0,...,20} {
			\draw[dotted] (\i,0) -- (\i,3) node [below] at (\i,0) {$\i$};
		}
		\foreach \i in {0,...,3} {
			\draw[dotted] (0,\i) -- (20,\i);
		}
		\node[minimum width=3\scaledx-0.05\scaledx,minimum height=1\scaledy-0.05\scaledy,inner sep=0,rectangle,draw,anchor=south west,pattern color=lightgray,pattern=north west lines] at (0,2) {\textbf{1}-\textbf{1}};
		\node[minimum width=3\scaledx-0.05\scaledx,minimum height=1\scaledy-0.05\scaledy,inner sep=0,rectangle,draw,anchor=south west,pattern color=lightgray,pattern=crosshatch] at (10,2) {\textbf{3}-\textbf{2}};
		\node[minimum width=6\scaledx-0.05\scaledx,minimum height=1\scaledy-0.05\scaledy,inner sep=0,rectangle,draw,anchor=south west,pattern color=lightgray,pattern=grid] at (4,2) {\textbf{2}-\textbf{2}};
		\node[minimum width=4\scaledx-0.05\scaledx,minimum height=1\scaledy-0.05\scaledy,inner sep=0,rectangle,draw,anchor=south west,pattern color=lightgray,pattern=grid] at (0,1) {\textbf{2}-\textbf{1}};
		\node[minimum width=3\scaledx-0.05\scaledx,minimum height=1\scaledy-0.05\scaledy,inner sep=0,rectangle,draw,anchor=south west,pattern color=lightgray,pattern=north west lines] at (4,1) {\textbf{1}-\textbf{2}};
		\node[minimum width=8\scaledx-0.05\scaledx,minimum height=1\scaledy-0.05\scaledy,inner sep=0,rectangle,draw,anchor=south west,pattern color=lightgray,pattern=crosshatch] at (13,1) {\textbf{3}-\textbf{3}};
		\node[minimum width=9\scaledx-0.05\scaledx,minimum height=1\scaledy-0.05\scaledy,inner sep=0,rectangle,draw,anchor=south west,pattern color=lightgray,pattern=crosshatch] at (0,0) {\textbf{3}-\textbf{1}};
		\node[minimum width=1\scaledx-0.05\scaledx,minimum height=1\scaledy-0.05\scaledy,inner sep=0,rectangle,draw,anchor=south west,pattern color=lightgray,pattern=north west lines] at (12,0) {\textbf{1}-\textbf{3}};
		\node[minimum width=2\scaledx-0.05\scaledx,minimum height=1\scaledy-0.05\scaledy,inner sep=0,rectangle,draw,anchor=south west,pattern color=lightgray,pattern=grid] at (10,0) {\textbf{2}-\textbf{3}};
		\draw[ultra thick,double] (9,0) -- (9,1);
		\draw[ultra thick,double] (7,1) -- (7,2);
		\draw[ultra thick,double] (10,2) -- (10,3);
	\end{tikzpicture}
	\figspace
	\caption{Decomposed schedule for the example JSP instance in Listing~\ref{prg:facts}\label{fig:decomposed}}
\end{figure}

In order to counteract limitations of window-wise successive optimization
due to ``decomposition mistakes'' as well as unnecessary idle times that do not directly
affect the makespan, 
we have devised two additional techniques that can be incorporated into the
iterative scheduling process.
The first extension is time window \emph{overlapping}, where a configurable percentage
of the operations per time window can still be rescheduled when proceeding
to the next time window.
To this end, an (optimized) partial schedule is postprocessed and the configured
number of operations to overlap are picked in decreasing order of starting times,
as encoded by the following stratified ASP program:
\begin{lstlisting}[numbers=none,frame=none,breaklines=false,fontadjust =true, basewidth= {0.58em, 0.45em}]
current((J,S),T,P,W) :- operation(J,S,M,P), start((J,S),T,W),
                        window(J,S,W).
current((J,S),T,P,W) :- operation(J,S,M,P), start((J,S),T,W),
                        overlap((J,S),W - 1).
	
inverse(O,N,W) :- current(O,T,P,W),
        N = #count{O' : current(O',T',P',W), (T,P,O) < (T',P',O')}.
overlap(O,W)   :- inverse(O,N,W), portion(C), N < C.
\end{lstlisting}
In addition to facts specifying a JSP instance and time windows
like in Listings~\ref{prg:facts} and~\ref{prg:tw}, the inputs
consist of reified facts of the forms
\lstinline{start(}$o$\lstinline{,}$t$\lstinline{,}$w$\lstinline{).}
and
\lstinline{overlap(}$o$\lstinline{,}$w-1$\lstinline{).}, giving the
starting times~$t$ of operations~$o$ in a partial schedule for time
window~$w$ along with the overlapped operations picked for the
previous time window $w-1$ (if any),
as well as a fact of the form
\lstinline{portion(}$c$\lstinline{).} whose argument~$c$ provides the
absolute number of operations to reschedule per time window.
Then, (derived) facts over \lstinline{current/4} indicate the
operations that were scheduled for the time window~$w$,
including those belonging to the overlap from time window $w-1$,
together with their starting and processing times.
The latter are considered to determine (derived) facts of the form
\lstinline{inverse(}$o$\lstinline{,}$n$\lstinline{,}$w$\lstinline{).}
such that indexes $n$ ranging from \lstinline{0}
are decreasing by starting times, where
processing times and operation identifiers serve as tie-breakers.
If $n<c$ applies, the respective operation~$o$ belongs to the $c$
operations with the latest starting times for the time window~$w$
and is taken as overlap represented in the form
\lstinline{overlap(}$o$\lstinline{,}$w$\lstinline{).}
For example, if one operation, matching 20\% of the size~$5$
of time windows,
from the first time window is taken as overlap for the
decomposed schedule in Figure~\ref{fig:decomposed}, the second operation of job~\lstinline{2}
with the latest starting time~\lstinline{4}
(and processing time~\lstinline{6}) is chosen in view of the
derived atoms
\lstinline{inverse((2,2),4,1)} and \lstinline{overlap((2,2),1)},
which then enables its processing after the second operation of job~\lstinline{3}
by rescheduling together with operations of the second time window.

The addition of overlapping operations to the encoding in Listing~\ref{prg:encoding}
requires handling them similar to operations of the time window to schedule,
i.e., enabling the choice of an execution order by the rules in lines
\ref{prg:base:order}-\ref{prg:base:order:end} rather than fixing the order
by the rule in line~\ref{prg:base:ordered}.
Given that overlapped operations~$o$ for a time window $w-1$ are 
supplied by reified facts of the form
\lstinline{overlap(}$o$\lstinline{,}$w-1$\lstinline{).},
augmenting the \lstinline{step(w)} subprogram in Listing~\ref{prg:encoding} with the rules
\begin{lstlisting}[numbers=none,frame=none,breaklines=false,lineskip=-0.5pt]
share((J1,S1),(J2,S2),P1,P2,0,w) :- operation(J1,S1,M,P1),
                                    operation(J2,S2,M,P2),
                                    overlap((J1,S1),w-1),
                                    window(J2,S2,w).
share((J1,S1),(J2,S2),P1,P2,1,w) :- operation(J1,S1,M,P1),
                                    operation(J2,S2,M,P2),
                                    overlap((J1,S1),w-2),
                                    window(J2,S2,w),
                                    not overlap((J1,S1),w-1).
share((J1,S1),(J2,S2),P1,P2,1,w) :- operation(J1,S1,M,P1),
                                    operation(J2,S2,M,P2),
                                    overlap((J1,S1),w-2),
                                    overlap((J2,S2),w-1),
                                    not overlap((J1,S1),w-1).
\end{lstlisting}
as well as the additional body literals
\lstinline{not overlap((J1,S1),w-1)} or 
\lstinline{not overlap(O,w-1)}, respectively,
for the rules in lines
\ref{prg:base:sharing_machine:begin}-\ref{prg:base:sharing_machine:end},
\ref{prg:diff_log:freez1:begin}, and
\ref{prg:diff_log:freez2:begin}
enables the rescheduling of overlapped operations together
with the next time window,
or fixes the execution order and starting times in case of
operations that were but are no longer overlapping.

\begin{lstlisting}[float=t,label=prg:compress,caption={Schedule compression encoding},lastline=52,xleftmargin=1.5em,fontadjust =true, basewidth= {0.58em, 0.45em}]
compress(J,S,M,P,T,W) :- operation(J,S,M,P), start((J,S),T,W), #(\label{prg:compress:compress1a}#)
                         window(J,S,W). #(\label{prg:compress:compress1b}#)
compress(J,S,M,P,T,W) :- operation(J,S,M,P), start((J,S),T,W), #(\label{prg:compress:compress2a}#)
                         overlap((J,S),W - 1). #(\label{prg:compress:compress2b}#)

constant(J,S,M,T,T + P,W) :- operation(J,S,M,P), start((J,S),T,W), #(\label{prg:compress:constant1a}#)
                             not compress(J,S,M,P,T,W). #(\label{prg:compress:constant1b}#)

traverse(J,S,M,P,W,I) :- compress(J,S,M,P,T,W), #(\label{prg:compress:traverse1a}#)
      I = #count{J',S': compress(J',S',M',P',T',W),(T',J')<(T,J)}. #(\label{prg:compress:traverse1b}#)
traverse(I)           :- traverse(J,S,M,P,W,I). #(\label{prg:compress:traverse2}#)

occupied(M,0,0,0,0) :- traverse(0), operation(J,S,M,P). #(\label{prg:compress:occupied1}#)
occupied(M,T,U,N,0) :- traverse(0), constant(J,S,M,T,U,W), #(\label{prg:compress:occupied2a}#)
           N = #count{J',S' : constant(J',S',M,T',U',W), T' <= T}. #(\label{prg:compress:occupied2b}#)
occupied(M,T,U,N,I) :- traverse(I), reinsert(J,S,M,T,U,W,N-1,I-1). #(\label{prg:compress:occupied3}#)
occupied(M,T,U,N,I) :- traverse(I), occupied(M,T,U,N',I - 1), #(\label{prg:compress:occupied4a}#)
                       reinsert(J,S,M,T',U',W,N'',I - 1),
		       D = N' - N'', E = (D + 1 - |D - 1|) / 2,
		       N = N' + (E + |E|) / 2. #(\label{prg:compress:occupied4b}#)
occupied(M,T,U,N,I) :- traverse(I), occupied(M,T,U,N,I - 1), #(\label{prg:compress:occupied5a}#)
                       reinsert(J,S,M',T',U',W,N',I - 1), M' != M. #(\label{prg:compress:occupied5b}#)

finished(M,N,0) :- occupied(M,T,U,N,0), #(\label{prg:compress:finished1a}#)
      N = #count{J',S' : constant(J',S',M,T',U',W)}. #(\label{prg:compress:finished1b}#)
finished(M,N,I) :- traverse(I), finished(M,N - 1,I - 1), #(\label{prg:compress:finished2a}#)
                   reinsert(J,S,M,T,U,W,N',I - 1). #(\label{prg:compress:finished2b}#)
finished(M,N,I) :- traverse(I), finished(M,N,I - 1), #(\label{prg:compress:finished3a}#)
                   reinsert(J,S,M',T,U,W,N',I - 1), M' != M. #(\label{prg:compress:finished3b}#)

released(J,1,M,P,0,W,I)       :- traverse(J,1,M,P,W,I). #(\label{prg:compress:released1}#)
released(J,S,M,P,T' + P',W,I) :- traverse(J,S,M,P,W,I), #(\label{prg:compress:released2a}#)
                                 start'((J,S - 1),T',W),
                                 operation(J,S - 1,M',P'). #(\label{prg:compress:released2b}#)

consider(J,S,M,P,R,T,U,W,0,I) :- released(J,S,M,P,R,W,I), #(\label{prg:compress:consider1a}#)
                                 occupied(M,T,U,0,I). #(\label{prg:compress:consider1b}#)
consider(J,S,M,P,R,T,U,W,N,I) :- consider(J,S,M,P,R,T',U',W,N-1,I), #(\label{prg:compress:consider2a}#)
                                 occupied(M,T,U,N,I),
				 T < (R + U' + |R - U'|) / 2 + P. #(\label{prg:compress:consider2b}#)

reinsert(J,S,M,R',R' + P,W,N,I) :- consider(J,S,M,P,R,T,U,W,N,I), #(\label{prg:compress:reinsert1a}#)
                                   finished(M,N,I),
			           R' = (R + U + |R - U|) / 2. #(\label{prg:compress:reinsert1b}#)
reinsert(J,S,M,R',R' + P,W,N,I) :- consider(J,S,M,P,R,T,U,W,N,I), #(\label{prg:compress:reinsert2a}#)
                                   occupied(M,T',U',N + 1,I),
				   R' = (R + U + |R - U|) / 2,
				   R' + P <= T'. #(\label{prg:compress:reinsert2b}#)

start'((J,S),T,W) :- constant(J,S,M,T,U,W). #(\label{prg:compress:start1}#)
start'((J,S),T,W) :- reinsert(J,S,M,T,U,W,N,I). #(\label{prg:compress:start2}#)

% #show compress/6.
% #show constant/6.
% #show traverse/6.
% #show occupied/5.
% #show finished/3.
% #show released/7.
% #show consider/10.
% #show reinsert/8.
#show start'/3.
\end{lstlisting}
%
As the second extension, we can make use of the stratified ASP program in Listing~\ref{prg:compress} to
postprocess an (optimized) partial schedule by inspecting operations of the
latest time window in the order of starting times whether idle slots
on their machines allow for an earlier execution.
To this end, the encoded greedy \emph{compression} approach
traverses the operations of the latest time window in (any
total) non-decreasing order of starting times and 
accomplishes the following step for each operation:
remove the operation from the schedule and reinsert it at the earliest time after the completion of its predecessor operation (if any) such that the machine executing the operation is available for the operation's processing time.
In the worst case, an operation is reinserted into the partial schedule at its original starting time,
so that the postprocessing never deteriorates the makespan.

In more detail, the rules in lines
\ref{prg:compress:compress1a}-\ref{prg:compress:constant1b}
distinguish the operations scheduled for the current time window
from those whose starting times have already been fixed before.
An order to traverse the former one by one according to their starting times is determined by the rules in lines
\ref{prg:compress:traverse1a}-\ref{prg:compress:traverse2}.
Beforehand, the time intervals $[t,u]$ at which machines~$m$
process previously scheduled operations are represented by atoms
of the form
\lstinline{occupied(}$m$\lstinline{,}$t$\lstinline{,}$u$\lstinline{,}$n$\lstinline{,}$0$\lstinline{)},
derived by the rules in lines
\ref{prg:compress:occupied1}-\ref{prg:compress:occupied2b},
where 
\lstinline{occupied(}$m$\lstinline{,}$0$\lstinline{,}$0$\lstinline{,}$0$\lstinline{,}$0$\lstinline{)}
is included as base case and the execution of the $n$-th operation
in the order of starting times on machine~$m$ is indicated otherwise.
The last interval~$n$ at which a machine~$m$ is processing some
previously scheduled operation is pointed out by an
\lstinline{finished(}$m$\lstinline{,}$n$\lstinline{,}$0$\lstinline{)}
atom provided by the rule in lines
\ref{prg:compress:finished1a}-\ref{prg:compress:finished1b}.
Moreover, the rules in lines
\ref{prg:compress:released1}-\ref{prg:compress:released2b}
yield atoms of the form
\lstinline{released(}$j$\lstinline{,}$s$\lstinline{,}$m$\lstinline{,}$p$\lstinline{,}$r$\lstinline{,}$w$\lstinline{,}$i$\lstinline{)},
where $(j,s)$ is traversed as $i$-th operation during compression and
$r$ is the completion time of its predecessor operation, which may itself result from a move to an earlier idle slot, or $0$ if $(j,1)$
is the first operation of its job~$j$.
The earliest starting time given by~$r$ is inspected by the rules
in lines
\ref{prg:compress:consider1a}-\ref{prg:compress:consider2b},
which iterate over the time intervals $[t,u]$ at which the machine~$m$ processing $(j,s)$ is occupied and stop with an atom
\lstinline{consider(}$j$\lstinline{,}$s$\lstinline{,}$m$\lstinline{,}$p$\lstinline{,}$r$\lstinline{,}$t$\lstinline{,}$u$\lstinline{,}$w$\lstinline{,}$n$\lstinline{,}$i$\lstinline{)}
at the first such $[t,u]$ interval after which $(j,s)$ can be reinserted into the partial schedule.
If this $n$-th time interval is the last at which machine~$m$ is processing some previously scheduled operation, the rule in lines
\ref{prg:compress:reinsert1a}-\ref{prg:compress:reinsert1b} applies,
or the rule in lines
\ref{prg:compress:reinsert2a}-\ref{prg:compress:reinsert2b}
otherwise,
and in either case $(j,s)$ is reinserted at the starting time
$r'=\max(\{r,u\})$, as expressed by an atom
\lstinline{reinsert(}$j$\lstinline{,}$s$\lstinline{,}$m$\lstinline{,}$r'$\lstinline{,}$r'+p$\lstinline{,}$w$\lstinline{,}$n$\lstinline{,}$i$\lstinline{)}.
The resulting partial schedule incorporates
$[r',r'+p]$ as the $(n+1)$-th time interval at which the machine~$m$
is occupied, where the rules in lines
\ref{prg:compress:occupied3}-\ref{prg:compress:occupied4b} and
\ref{prg:compress:finished2a}-\ref{prg:compress:finished2b}
update the intervals for $m$,
while the rules in lines
\ref{prg:compress:occupied5a}-\ref{prg:compress:occupied5b} and
\ref{prg:compress:finished3a}-\ref{prg:compress:finished3b}
reflect that the operations processed by other machines remain unchanged.
Finally, the rules in lines 
\ref{prg:compress:start1}-\ref{prg:compress:start2} provide
the compressed partial schedule in terms of atoms of the form
\lstinline{start'((}$j$\lstinline{,}$s$\lstinline{),}$t$\lstinline{,}$w$\lstinline{)}, thus resembling the input facts
\lstinline{start((}$j$\lstinline{,}$s$\lstinline{),}$t$\lstinline{,}$w$\lstinline{).}

As mentioned above, the updated starting times resulting from the
compression are earlier or the same as the original starting times calculated in the actual makespan optimization.
Hence, the latter can be safely exchanged by mapping
\lstinline{start'(}$o$\lstinline{,}$t$\lstinline{,}$w$\lstinline{)}
atoms back to
\lstinline{start(}$o$\lstinline{,}$t$\lstinline{,}$w$\lstinline{)}
for proceeding with the iterative scheduling process, i.e.,
determining operations to overlap and moving on to the next time
window.
For example, when compressing the decomposed schedule 
in Figure~\ref{fig:decomposed},
the starting time~\lstinline{12} of the third operation of job~\lstinline{1}
is turned into~\lstinline{9} to fill the idle slot available on
machine~\lstinline{3}, which is reflected by 
\lstinline{start'((1,3),9,2)} in the answer set obtained with the
encoding in Listing~\ref{prg:compress}.
Then, \lstinline{start((1,3),12,2)} can be replaced by
\lstinline{start((1,3),9,2)} to make machine~\lstinline{3}
available from time~$12$ instead of~$13$ in case there were a
third time window with operations to be processed by machine~\lstinline{3}.

\section{Experiments}
\label{sec:experiments}
For evaluating our multi-shot ASP modulo DL approach to JSP solving,
we ran experiments on JSP benchmark sets due to Taillard
and Demirkol \cite{taillard1993benchmarks,demirkol1998benchmarks},
each including ten instances with $50\times15$ jobs and machines.
The instances are generated such that each job consists of $15$ operations, where the sequence of machines processing the operations varies from job to job.
\addtext{In \cite{elscge22a}, we focussed on these instances as well because they yield
representative relative results among different decomposition strategies, while their
uniform size helps for configuring an appropriate number of time windows to use with every
decomposition strategy.
For scalability experiments in the second part of this section,
we additionally run selected decomposition strategies on a benchmark set of industrial-size JSP instances
due to Da Col and Teppan \cite{da2022industrial}.}

\begin{table}[t]
	\caption{Overview of our experimental evaluation\label{tab:comparisons}}
	\setlength{\tabcolsep}{0.5pt}
	\centering
	\begin{tabular}{l c c}
		\toprule
		Question & Reference(s) & Table \\
		\midrule
		Which number/size of time windows is adequate? & \cite{el2022problem} & \ref{tab:Table01} \\
		Does the compression technique improve the solution quality? & \cite{el2022problem} & \ref{tab:Table01} \\
		Which static decomposition strategies improve the solution quality? & \cite{el2022problem,elscge22a} & \ref{tab:Table02} \\
		Can problem decomposition benefit from heuristic search methods? & & \ref{tab:Table03} \\
		Can overlapping techniques further improve the solution quality? & \cite{el2022problem} & \ref{tab:Table04} \\
		How does \clingodl\ compare to state-of-the-art CP systems? & \cite{el2022problem} & \ref{tab:Table05} \\
		How do multi-shot ASP modulo DL solving and CP systems scale? & & \ref{tab:Table06} \\
		\bottomrule
	\end{tabular}
\end{table}

In our experiments,%
\footnote{The JSP benchmark sets and our implementation are available at: \url{https://github.com/prosysscience/Job-Shop-Scheduling}}
we assess the decomposition strategies presented in Sections~\ref{subsec:decomposition} 
and~\ref{subsec:method}
with different numbers (and sizes) of time windows as well as the impact of the time window overlapping and compression techniques described in Section~\ref{subsec:scheduling}.
For the comparability of results between runs with a different number of time windows
and respective optimization subproblems addressed by multi-shot solving,
we divide the total runtime limit of $1000$ seconds for \clingodl (version 1.3.0) by the number
of time windows to evenly spend optimization efforts on subproblems,
where all experiments have been run on an 
Intel\textsuperscript{\textregistered} Core\texttrademark{} i7-8650U CPU 
Dell Latitude 5590 machine under Windows 10.
\revisedtext{%
The main questions that we address in the following are outlined in Table~\ref{tab:comparisons},
indicating which of our studies \cite{el2022problem,elscge22a}
also provides corresponding results as well as the respective table in this paper.
Notably, we below arrive at a different outcome regarding overlapping techniques (Table~\ref{tab:Table04}) than \cite{el2022problem}, due to turning
to heuristic schedules rather than static properties for problem decomposition.}%
\begin{table}[t] 
	\caption{Comparison of numbers of time windows without/with compression using M-EST decomposition\label{tab:Table01}} 
	\setlength{\tabcolsep}{6pt}
	\centering
	\begin{tabular}{r l r r r r r r r}
		\toprule
		Instances & Average & 1 & 2 & 3 & 4 & 5 & 6 & 10  \\
		\midrule
		Taillard       & Makespan             & $3542.1$   & $3149.7$	     &  \boldmath{$3083.8$} & $3110.7$ & $3213.6$ & $3225.3$ & $3524.9$ \\
		& Time             & $1000.0$	& $1000.0$   & $1000.0$	 & $934.4$ & $615.5$  & $243.8$  & $18.5$  \\
		& Interrupts    & $1.0$	    & $2.0$      & $3.0$	& $3.6$  & $2.7$  & $1.0$  & $0.1$       \\ [1.5mm]
		
		Demirkol       & Makespan             & $8589.8$   & $7436.1$	      & \boldmath{$7283.2$}  &  $7327.8$ &  $7548.0$ & $7727.4$ & $8912.7$ \\
		& Time             & $1000.0$	    & $1000.0$	   & $1000.0$	  & $882.4$  & $640.2$  & $486.8$   & $68.7$     \\
		& Interrupts    & $1.0$	    & $2.0$      & $3.0$	& $3.4$ & $2.8$  & $2.4$  & $0.5$      \\
		\midrule
		Taillard       & Makespan             & $3542.1$   & $3068.6$	     &  $3015.9$ & \boldmath{$3010.3$} & $3040.7$ & $3058.2$ & $3151.6$ \\
		& Time             & $1000.0$	& $1000.0$   & $1000.0$	 & $907.5$ & $573.8$  & $274.3$  & $18.6$  \\
		& Interrupts    & $1.0$	    & $2.0$      & $3.0$	& $3.4$  & $2.5$  & $1.4$  & $0.1$       \\ [1.5mm]
		
		Demirkol       & Makespan             & $8589.8$   & $7138.2$	      & $6734.4$  &  $6727.2$ &  \boldmath{$6698.6$} & $6840.9$ & $7063.4$ \\
		& Time             & $1000.0$	    & $1000.0$	   & $970.5$	  & $826.9$  & $541.5$  & $307.8$   & $35.0$     \\
		& Interrupts    & $1.0$	    & $2.0$      & $2.9$	& $3.1$ & $2.2$  & $1.4$  & $0.1$      \\
		\bottomrule
	\end{tabular}
\end{table}

Our first comparison, shown in Table~\ref{tab:Table01}, concerns
the Machine-based Earliest Starting Time (M-EST) decomposition
strategy, which turned out as most effective among the
considered static decomposition strategies, with varying numbers
of time windows in separate columns and compression excluded in the
upper or included in the lower part, respectively.
For both benchmark sets, we report the average makespan over the ten instances, where smaller values indicate better schedules, the average runtime of \clingodl, and the average number of interrupted optimization processes on subproblems,
where the best partial schedule obtained in time is taken to progress
with the iterative scheduling.
We gradually increase the number of time windows from~$1$ to~$6$ and additionally include results for~$10$ time windows to outline the trend of degrading solution quality when the partition into time windows becomes too fragmented.
In fact, the shortest average makespans, highlighted in boldface, are obtained with problem decomposition into $3$, $4$, or $5$ time windows\addtext{, each consisting of $750/3=250$, $750/4\approx 187$, or $750/5=150$ operations, respectively}.
We observe that the average makespans with compression in the lower
part of Table~\ref{tab:Table01} are substantially shorter, apart from
the indifferent single-shot optimization with $1$ time window only.
The latter represents global optimization on the full problem,
and decompositions into more than one time window clearly improve the solution quality.
These advantages are not surprising, considering that the JSP instances are highly combinatorial \cite{shysha18a} and each global optimization run times out with a more or less optimized solution.
Since the compression technique consistently improves the solution quality, we commit to it in the comparisons addressed below.
We also fix the number of time windows to $3$ from now on in order
to reduce variable parameters, while other numbers may occasionally be better.
\begin{table}[b] 
	\caption{Comparison of different static decomposition strategies\label{tab:Table02}}
	\setlength{\tabcolsep}{6.8pt}
	\centering
	\begin{tabular}{r r r r r r r }
		\toprule
		& \multicolumn{3}{c}{Taillard} & \multicolumn{3}{c}{Demirkol} \\
		Strategy & Makespan & Time & Interrupts & Makespan & Time & Interrupts \\
		\midrule
		J-EST            & $3492.1$ & $1000.0$   & $3.0$	   & $7789.3$      & $1000.0$ & $3.0$ \\ 
		J-MTWR            & $3440.2$  & $1000.0$   & $3.0$	& $7738.2$      & $1000.0$ & $3.0$  \\ 
		[1.5mm]
		M-EST             & \boldmath{$3015.9$}  & $1000.0$   & $3.0$	& \boldmath{$6734.4$}      & $970.5$ & $2.9$ \\
		M-MTWR        & $3060.3$  & $1000.0$   & $3.0$	& $6861.5$      & $1000.0$ & $3.0$ \\ 
		[1.5mm]
		Clustering            & $3256.4$  & $1000.0$   & $3.0$	& $7174.6$     & $1000.0$ & $3.0$ \\
		\bottomrule
	\end{tabular}
\end{table}

In Table~\ref{tab:Table02}, we compare the previously considered
M-EST decomposition to the Machine-based Most Total Work Remaining
(M-MTWR) strategy, their Job-based versions J-EST and J-MTWR, as
well as constrained clustering using the best feature set among those
investigated in~\cite{elscge22a} for each instance.
The considerably increased average makespans in the first two rows clearly indicate that
time windows determined with Job-based decomposition strategies are less adequate than
those investigating bottleneck machines in the first place, and then picking their operations
based on the smallest EST or greatest MTWR value as a secondary criterion.
While problem decomposition by constrained clustering also improves
over the Job-based J-EST and J-MTWR strategies, it remains behind
M-EST and M-MTWR, whose differences in average makespan are less
pronounced. 
This observation suggests that bottleneck machines should play the most prominent role in decomposition and scheduling heuristics, while other features provide secondary criteria for picking operations.

While the decomposition strategies considered so far were based on static properties, Table~\ref{tab:Table03} provides results for
decomposition 
based on the starting times of operations
in schedules obtained with greedy search methods using
\textit{First-In-First-Out} (FIFO), \textit{Most-Total-Work-Remaining} (MTWR), and \textit{Reinforcement Learning} (RL) \cite{tassel2021reinforcement}
\addtext{%
as heuristics for selecting the next operation to allocate
in step~2 of Figure~\ref{fig:greedy}.
The job features explored by the RL policy, namely, the remaining
processing time (\textit{RPT}) and the waiting time (\textit{WT}),
combine the FIFO and MTWR criteria for selecting operations.
Going beyond the basic FIFO and MTWR heuristics,
the RL policy includes flexibility to perform noops in step~3
of Figure~\ref{fig:greedy}, meaning that a machine can be kept
free for operations getting released later on, which enables global solutions that are less greedy than operation allocation by FIFO or MTWR.}
We observe that the average makespan of schedules for FIFO and MTWR
is larger than with the best static strategy M-EST,
while the latter is outperformed by RL.
Hence, we conclude that a machine learning approach should be used
to generate initial schedules, which can then be decomposed
\addtext{%
into time windows by taking operations in the order of their starting times and further optimizing the time windows one after the other}.
Moreover, our preliminary experiments with static decomposition strategies suggested that overlapping techniques can be beneficial,
yet we obtain negative results for the RL decomposition in Table~\ref{tab:Table04}.
\addtext{%
In fact, the RL policy is (re)trained on each instance and then manages
to generate a good-quality schedule that is all but easy to improve further, which is a likely reason that the possibility to reschedule overlapping operations does not yield solutions of better quality.}
\begin{table}[t] 
	\caption{Comparison of FIFO, MTWR, and RL decomposition\label{tab:Table03}}
	\setlength{\tabcolsep}{6.8pt}
	\centering
	\begin{tabular}{r r r r r r r }
		\toprule
		& \multicolumn{3}{c}{Taillard} & \multicolumn{3}{c}{Demirkol} \\
		Strategy & Makespan & Time & Interrupts & Makespan & Time & Interrupts \\
		\midrule
		FIFO            & $3257.1$ & $1000.0$   & $3.0$	   & $7880.1$      & $1000.0$ & $3.0$ \\ 
		MTWR             & $3234.1$  & $1000.0$   & $3.0$	& $7109.6$      & $1000.0$ & $3.0$ \\
		RL            & \boldmath{$2943.5$}  & $1000.0$   & $3.0$	& \boldmath{$6458.3$}     & $1000.0$ & $3.0$ \\
		\bottomrule
	\end{tabular}
\end{table}

\begin{table}[b] 
	\caption{Comparison of time window overlapping techniques using RL decomposition\label{tab:Table04}}
	\setlength{\tabcolsep}{10.1pt}
	\centering
	\begin{tabular}{r r r r r r r}
		\toprule
		& \multicolumn{3}{c}{Taillard} & \multicolumn{3}{c}{Demirkol} \\
		Overlap & Makespan & Time & Interrupts & Makespan & Time & Interrupts  \\
		\midrule
		0\%              & \boldmath{$2943.5$}  & $1000.0$   & $3.0$	& \boldmath{$6458.3$}      & $1000.0$ & $3.0$ \\
		
		10\%             & $2952.9$  & $1000.0$   & $3.0$	& $6486.9$     & $1000.0$ & $3.0$ \\ 
		
		20\%             & $2954.9$ & $1000.0$   & $3.0$	   & $6563.6$      & $1000.0$ & $3.0$ \\ 
		30\%             & $2969.0$  & $1000.0$   & $3.0$	& $6621.7$      & $1000.0$ & $3.0$  \\ 
		\bottomrule
	\end{tabular}
\end{table}

\addarea{%
\begin{table}[t]\addcolor
	\caption{\addcolor{}Comparison of single- and multi-shot ASP modulo DL solving approaches to CP-Optimizer and OR-Tools\label{tab:Table05}}
	\setlength{\tabcolsep}{7.5pt}
	\centering
	\begin{tabular}{r r r r r}
	\toprule
	& \multicolumn{2}{c}{Taillard} & \multicolumn{2}{c}{Demirkol}
	\\
	Strategy & Makespan & Distance & Makespan & Distance
	\\\midrule
	Optima & $2773.8$ & $0.0$ & $5894.3$--$5950.6$ & $0.0$ 
	\\[1.5mm]
	CP-Optimizer & \boldmath{$2773.8$} & $0.0$ & \boldmath{$6211.5$} & $260.9$--$317.2$
	\\
	OR-Tools & $2778.3$ & $4.5$ & $6293.3$ & $342.7$--$399.0$
	\\[1.5mm]
	Multi-shot & $2943.5$ & $169.7$ & $6458.3$ & $507.7$--$564.0$
	\\
	Single-shot & $3542.1$ & $768.3$ & $8589.8$ & $2639.6$--$2695.5$
	\\
	\bottomrule
	\end{tabular}
\end{table}}

In Table~\ref{tab:Table05}, we additionally compare the average makespan of
schedules obtained with single- and multi-shot ASP modulo DL solving,
the latter taking RL solutions for problem decomposition, 
to the (global) solution quality achieved by the state-of-the-art CP systems
\addtext{%
CP-Optimizer (version 22.1.1) \cite{laroshvi18a} and OR-Tools (version 9.5) \cite{pediga23a}.
We use 
CP encodings of JSP supplied by the CP system developers
\cite{cpoptimizer,ortools},
set the runtime limit to $1000$ seconds,
as also taken for \clingodl,
and indicate the distance between the average makespan of obtained schedules and the optima known for Taillard's or within
ranges for Demirkol's instances.
We observe that CP-Optimizer,
which succeeds to find optimal schedules and terminates its runs before the time limit
for Taillard's instances, as well as OR-Tools have an edge on our single- and multi-shot ASP modulo DL
approaches.
This can be explained by the availability of interval variables and global constraints in the CP encodings, which turn out to be particularly 
effective in modeling the non-intersecting execution of operations 
on each machine.
However,
the RL decomposition and multi-shot optimization
help \clingodl{}
to come substantially closer than single-shot ASP modulo DL solving on the full problem.}

\addarea{%
The impressive performance of CP systems for JSP solving is also 
confirmed by the experimental observations made in \cite{da2022industrial,tassel2021reinforcement}, and the JSP instances with $50\times15$ jobs and machines are small enough for them to find good-quality or even optimal solutions in limited time.
For further assessing the scaling behavior,
we extend the scope to industrial-size instances due to Da Col and Teppan \cite{da2022industrial} with $100\times100$ jobs and machines.
As these instances were not considered in \cite{tassel2021reinforcement},
initial schedules of the RL policy are unavailable, and we apply the
J-EST, J-MTWR, M-EST, and M-MTWR strategies for problem decomposition
into $30$ time windows comprising $10000/30 \approx 333$ operations each.
For reference, Table~\ref{tab:Table06} reports the average makespans
obtained by running the single-core configurations of CP-Optimizer and OR-Tools for $6$ hours on a 2 GHz AMD EPYC 7551P 32 Cores CPU machine
\cite{da2022industrial},
where CP-Optimizer has a significant edge on OR-Tools thanks to its
memory-efficient implementation of the global constraint establishing
non-intersecting operation execution for machines.%
\begin{table}[b]\addcolor
	\caption{\addcolor{}Comparison of different static decomposition strategies to CP-Optimizer and OR-Tools\label{tab:Table06}}
	\setlength{\tabcolsep}{18.2pt}
	\centering
	\begin{tabular}{r r r}
		\toprule
		& \multicolumn{2}{c}{Da Col \& Teppan}
		\\
		Strategy & Makespan & Time \\
		\midrule
		CP-Optimizer & \boldmath{$80565.5$} & $21600.0$ \\
		OR-Tools & $114476.2$ & $21600.0$ \\[1.5mm]
		J-EST & $103260.0$ & $10615.8$ \\
		J-MTWR & $100120.9$ & $10464.8$ \\[1.5mm]
		M-EST & $105925.9$ & $10327.4$ \\
		M-MTWR & $100793.2$ & $10503.9$ \\
		\bottomrule
	\end{tabular}
\end{table}

Accordingly, we also impose a total runtime limit of $6$ hours,
divided into up to $21600/30=720$ seconds per time window.
As a general trend, we see that the solution quality achieved by our
decomposition strategies lies in-between CP-Optimizer and OR-Tools,
with some advantages for the Job- and Machine-based MTWR strategies.
That is, the previously encountered gap to OR-Tools on moderately sized
instances is made up by better multi-shot optimization performance
of \clingodl{} on the downscaled subproblems.
The latter can also be observed on the average runtimes given in the third
column of Table~\ref{tab:Table06}, indicating that all decomposition strategies lead to optimal partial schedules for more than half of the
time windows and take only about $3$ of the $6$ hours available in total.
Clearly, this performance improvement comes to the cost of a reduced overall solution quality, as the decomposition into $30$ time windows constrains the potential execution order of operations.
Hence, the good-quality schedules found by CP-Optimizer remain unmatched
for the large-scale instances due to Da Col and Teppan.}


\section{Conclusions} \label{sec:conclusions}
Our work develops multi-shot ASP modulo DL methods for JSP solving by means of problem
decomposition into balanced time windows, which respect the operation precedence
within jobs, along with successive makespan minimization for extending partial schedules to a global solution.
\addtext{%
Several simple static decomposition strategies
(denoted by J-EST, J-MTWR, M-EST, and M-MTWR) can be directly
encoded in ASP, which provides declarative means to split
highly combinatorial JSP instances into better manageable subproblems,}
leading to substantially better schedules than
tackling the full problem at once with
single-shot ASP modulo DL optimization.
Our experiments with different decomposition strategies show that
(i) the consideration of bottleneck machines is beneficial when problem decomposition is based on simple static properties,
(ii) performing compression after optimizing each time window improves the quality of global solutions,
(iii) decomposing and further optimizing schedules generated by
greedy search methods is preferable to static decomposition strategies, even when the latter incorporate clustering
w.r.t.\ a variety of features,
and
(iv) time window overlapping does not improve the solution quality
further when the problem decomposition process incorporates proficient heuristic methods,
where 
RL policies turn out to be particularly successful.
\addtext{%
No matter which decomposition strategy is applied,
for runtime limits of $5$ to $12$ minutes per time window,
a few hundred operations in each time window,
namely, between $250$ and $350$ operations,
turned out to be adequate for achieving
a robust performance of optimizing partial schedules by \clingodl{}.}

\addtext{%
The main focus of our work is on investigating how problem decomposition
and multi-shot optimization can be exploited to reliably find good-quality
solutions for JSP instances on which single-shot ASP modulo DL solving fails.
Simple static decomposition strategies, especially the four we encoded in ASP, should thus be understood as tools for partitioning the operations
of JSP instances into time windows, yet without claiming that these
decomposition strategies would be very elaborate or highly likely
to come close to global optima.
The experimental finding that the RL decomposition performs best indicates that applying heuristic search methods to generate initial solutions that can then be optimized further is a promising approach,
which matches the general principle of Large Neighborhood Search \cite{pisrop10a}.}

The comparison with the state-of-the-art CP systems
\addtext{CP-Optimizer and} OR-Tools yields that
global optimization by the latter is still ahead of our multi-shot ASP modulo DL methods on \addtext{moderately sized JSP instances including $750$ operations to be scheduled}.
In fact, \addtext{CP-Optimizer and} OR-Tools take advantage of dedicated interval variables and global constraints, designed specifically for scheduling problems, in their CP encodings.
\addtext{%
However, for industrial-size instances with $10000$ operations,
OR-Tools does not scale well anymore, and the problem decomposition
promotes our multi-shot ASP modulo DL solving approach,
while the memory-efficient global constraint implementation
of CP-Optimizer remains unmatched.
We conclude that modeling scheduling problems in terms of the
global constraints supplied by CP systems,
which also outperform Mixed Integer Programming approaches \cite{kubec19a},
is likely to yield better single-shot optimization performance than the
plain difference constraints of \clingodl{}.
The first-order modeling language and multi-shot solving support
available in ASP modulo DL are nonetheless advantageous
for prototyping a variety of decomposition strategies,
while adopting CP systems for the window-wise successive optimization
of partial schedules would require dedicated implementations
that are beyond the scope of this paper.}

\revisedtext{Optima for highly combinatorial JSP instances can only be  
guaranteed in exceptional cases to which theoretical lower bounds apply \cite{taillard1993benchmarks}, while exhaustively traversing the search space with exact methods is beyond reach.
When complete optimization methods are interrupted at a given time limit,
they do not guarantee \emph{bounded suboptimality} of the best solution
obtained in time, i.e., the distance to an optimum is not bound by any certain range.
This situation does not change with our decomposition strategies, where the impact of ``decomposition mistakes'' on the makespan is difficult to predict.
Approximation methods that guarantee bounded suboptimality for JSP have attracted theoretical interest \cite{gopasrsw01a},
and approximations of bounded suboptimality have also been devised for the related problem of Multi-Agent Path Finding \cite{chligahastko22a}.
However, the latter are so far limited to step-wise agent moves, while JSP with its
non-uniform processing times comes closer to Multi-Agent Path Finding with continuous time \cite{anyasuatst22a}.}

\revisedtext{Regardless of (un)available guarantees, approximation methods, aiming to reduce the runtime for finding good-quality rather than (provably) optimal solutions, are of great practical relevance for scheduling problems beyond JSP and for ASP solving in general.
For instance, the scheduling of realistic semiconductor manufacturing processes involves jobs with hundreds of operations and large-scale factories with
thousands of machines, which are subject to
diverse conditions like (partially) flexible machine allocation,
preemptive maintenance, sequence-dependent setups, as well as
batching and cascading operation execution \cite{kohakamo20a}.
As a consequence, the scalability of exact ASP modulo DL solving approaches \cite{alelge23a,elalge23a} is limited to fragments of the full semiconductor manufacturing setting,
such as machine groups with specific functionalities,
while a local search technique based on ant colony optimization \cite{ali2024greedy} has been designed for global scheduling.}

\revisedtext{To handle their size and complexity, real-world optimization problems frequently call for multi-shot ASP solving methods.
Such problems include aircraft routing and maintenance planning \cite{tasrba21a},
where the turnaround times of aircrafts are gradually increased to achieve better anytime performance than exhaustive single-shot optimization taking the full problem upfront.
In \cite{caganoro23a}, Benders decomposition is applied to interleave solving the master problem of assigning medical treatments to days with the day-wise subproblem of scheduling the times and doctors to perform them.
Recent frameworks provide functionalities for Large Neighborhood Search in ASP (modulo DL) solving \cite{eigehimuoest24a,suinnascsotaba24b},
where parts of an initial solution are unassigned (``destroyed'') and then reassigned differently (``repaired'') in order to improve the solution quality.
While the idea is comparable to decomposing initial solutions obtained by FIFO, MTWR, or RL dispatching heuristics, Large Neighborhood Search is applied to the full ground instantiation of an optimization problem, without incorporating a concept like time windows to control the representation size for subproblems.}

\revisedtext{%
Clearly, the decomposition of large-scale instances into time windows is
specific to JSP, may be adopted to closely related scheduling problems,
such as flow-shop, open-shop, or flexible job-shop scheduling
\cite{taillard1993benchmarks,xigapelili19a},
but does not carry forward to other kinds of combinatorial optimization problems,
e.g., the well-known Knapsack Problem and the Traveling Salesperson Problem.
While structural approaches like tree decomposition \cite{fichte2017bounded} are domain-independent, their applicability relies on the features of problem instances, whose large treewidth may render tree decomposition ineffective.
Hence, the rapid prototyping and practical application of dedicated
problem decomposition and multi-shot ASP (modulo DL) solving methods
could be further promoted by future work supplying modeling directives similar to the
neighborhood declarations for ASP-based Large Neighborhood Search or
the domain-specific \lstinline{#heuristic} statements of \clingo \cite{gebser2013domain}.}

\paragraph{Acknowledgments.}
  This work was partially funded by KWF project 28472, cms electronics GmbH, FunderMax GmbH, Hirsch Armb\"ander GmbH, incubed IT GmbH, Infineon Technologies Austria AG, Isovolta AG, Kostwein Holding GmbH, and Privatstiftung K\"arntner Sparkasse.
  We are grateful to the anonymous reviewers for constructive 
  comments that helped to improve this paper.


	
	
	
	


\newcommand{\etalchar}[1]{$^{#1}$}

\end{document}